\documentclass{article}

\usepackage[final, nonatbib]{neurips_2024}

\usepackage[utf8]{inputenc} 
\usepackage[T1]{fontenc}    
\usepackage{hyperref}       
\usepackage{url}            
\usepackage{booktabs}       
\usepackage{amsfonts}       
\usepackage{nicefrac}       
\usepackage{microtype}      
\usepackage{xcolor}         
\usepackage{graphicx}
\usepackage{amsmath}
\usepackage{colortbl}
\usepackage{soul}
\usepackage{color}
\usepackage{wrapfig}
\usepackage{threeparttable}
\usepackage{caption}

\title{Grid4D: 4D Decomposed Hash Encoding for High-Fidelity Dynamic Gaussian Splatting}

\author{
Jiawei Xu\renewcommand{\thefootnote}{\arabic{footnote}}\footnotemark[1]\ ~\renewcommand{\thefootnote}{\fnsymbol{footnote}}\footnotemark[2]\ , \ 
Zexin Fan\renewcommand{\thefootnote}{\arabic{footnote}}\footnotemark[1]\ ~\renewcommand{\thefootnote}{\fnsymbol{footnote}}\footnotemark[2]\ , \ 
Jian Yang\renewcommand{\thefootnote}{\arabic{footnote}}\footnotemark[1]\ ~\renewcommand{\thefootnote}{\fnsymbol{footnote}}\footnotemark[4]~\ \footnotemark[1]\ , \
Jin Xie\renewcommand{\thefootnote}{\arabic{footnote}}\footnotemark[2]\ ~\footnotemark[3]\ ~\renewcommand{\thefootnote}{\fnsymbol{footnote}}\footnotemark[3]\ ~\footnotemark[1] \\
\renewcommand{\thefootnote}{\arabic{footnote}}\footnotemark[1] \ PCA Lab, VCIP, College of Computer Science, Nankai University \\
\renewcommand{\thefootnote}{\arabic{footnote}}\footnotemark[2] \ State Key Laboratory for Novel Software Technology, Nanjing University, Nanjing, China \\
\renewcommand{\thefootnote}{\arabic{footnote}}\footnotemark[3] \ School of Intelligence Science and Technology, Nanjing University, Suzhou, China \\
\footnotemark[2] \ \{jiaweixu, zexin\_fan\}@mail.nankai.edu.cn \ 
\footnotemark[3] \ csjxie@nju.edu.cn
\footnotemark[4] \ csjyang@nankai.edu.cn
}

\begin{document}

\maketitle

\renewcommand{\thefootnote}{\fnsymbol{footnote}}
\footnotetext[1]{Corresponding authors.}

\begin{abstract}

Recently, Gaussian splatting has received more and more attention in the field of static scene rendering.
Due to the low computational overhead and inherent flexibility of explicit representations, plane-based explicit methods are popular ways to predict deformations for Gaussian-based dynamic scene rendering models.
However, plane-based methods rely on the inappropriate low-rank assumption and excessively decompose the space-time 4D encoding, resulting in overmuch feature overlap and unsatisfactory rendering quality.
To tackle these problems, we propose Grid4D, a dynamic scene rendering model based on Gaussian splatting and employing a novel explicit encoding method for the 4D input through the hash encoding.
Different from plane-based explicit representations, we decompose the 4D encoding into one spatial and three temporal 3D hash encodings without the low-rank assumption.
Additionally, we design a novel attention module that generates the attention scores in a directional range to aggregate the spatial and temporal features.
The directional attention enables Grid4D to more accurately fit the diverse deformations across distinct scene components based on the spatial encoded features.
Moreover, to mitigate the inherent lack of smoothness in explicit representation methods, we introduce a smooth regularization term that keeps our model from the chaos of deformation prediction.
Our experiments demonstrate that Grid4D significantly outperforms the state-of-the-art models in visual quality and rendering speed.
Project page: ~\url{https://jiaweixu8.github.io/Grid4D-web/}.

\end{abstract}

\section{Introduction}\label{sec:introduction}

Dynamic scene rendering aims to construct dynamic scenes from images with specific camera poses and timestamps, allowing rendering from arbitrary viewpoints and moments.
Traditional methods use Neural Radiance Field~(NeRF)~\cite{nerf} and deformation fields to reconstruct dynamic scenes for arbitrary rendering.
However, these works rely on predicting deformations with the over-smooth full Multilayer Perceptron~(MLP)~\cite{d-nerf, fsdnerf, banmo, dynamic-nerf, dynibar, hypernerf, ndf, nerf-ds, dynerf, nerflow, li2021neural, ndr, nr-nerf, rodynrf, s-nerf, saff, 4dregsdf, star}, resulting in slow training speeds and artifacts in rendering quality.
To address these challenges, explicit representations such as planes~\cite{eg3d} and hash encoding~\cite{instant-ngp} have been introduced to enhance the rendering of dynamic scenes~\cite{tensor4d, k-planes, hexplane, 4k4d, tineuvox, masked-spacetime-hashing, forwardflowdnerf, factorized-motion, nerfplayer}.
The explicit representations store the intermediate features generated by the partial forward propagation process in a grid-like format.
This approach allows us to obtain intermediate features by directly interpolating the cached features based on the input, bypassing the need for the full forward propagation process.
In addition to reducing computing resource consumption, the inherent flexibility of explicit representation offers advantages in rendering more complex scenes.

Recently, Gaussian splatting~\cite{gaussian-splatting} achieved fast and high-fidelity rendering of static scenes.
Additionally, many works have employed Gaussian splatting for dynamic scene rendering by deforming Gaussians based on the timestamp~\cite{4d-gs, deformable-3d-gaussians, gaussian-flow, spacetime-gaussians, dynmf, md-splatting, dynamic-3d-gaussians, fudan4dgs, beida4dgs, gaufre, gags, sc-gs, cogs, 3dgstream}.
Deforming Gaussians through pre-defined functions is an effective way to reconstruct dynamic scenes with sufficient viewpoints~\cite{gaussian-flow, spacetime-gaussians}.
Additionally, implicit and explicit neural networks are more popular for deforming Gaussians in general cases~\cite{4d-gs, deformable-3d-gaussians, sc-gs}.
However, fully MLP-based implicit neural networks have limited learning capacity because of their over-smooth inherent property, thereby struggling to render several complex scenes and details effectively.
Hence, explicit representation might be an available method to address these problems.
Prior works such as 4D-GS~\cite{4d-gs} use plane-based explicit representations to predict Gaussian deformations, decomposing the 4D space-time encoding into a format comprising six 2D planes, but the performance remains unsatisfactory.
We consider that the plane-based methods for Gaussian deformation prediction are based on the low-rank assumption which assumes that the features for the deformations have a great deal of commonality and could be factorized into a very low-rank format~\cite{tensor4d, k-planes, hexplane, 4d-gs}.
As shown in Figure~\ref{fig:encoder}, when facing Gaussians with massive overlapping coordinates, the over-decomposition makes the features have excessive overlap which limits their discriminability for deformation prediction.
Therefore, such overlap might block the model from predicting different deformations, resulting in low rendering quality.

\begin{figure}[t]
  \centering
  \includegraphics[scale=0.52, trim=0 310 220 0, clip]{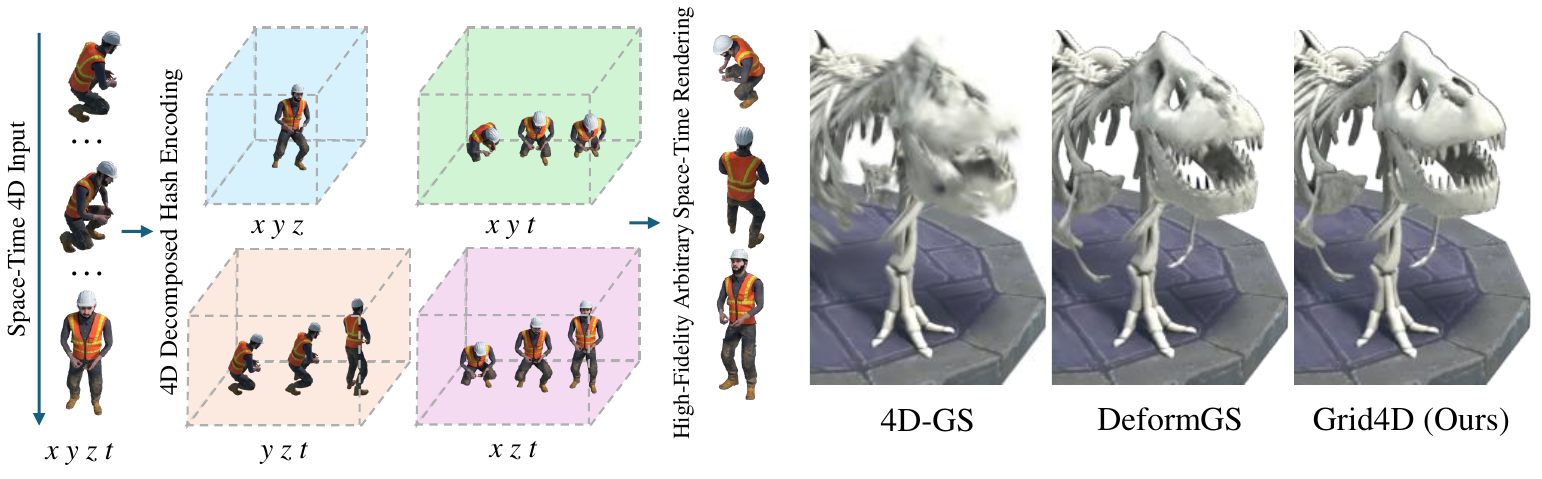}
  \caption{We propose a novel explicit representation method for dynamic scene rendering that decomposes the space-time 4D encoding into the 3D format without the unsuitable low-rank assumption. We achieve significant improvements over the state-of-the-art models~\cite{4d-gs, deformable-3d-gaussians} in rendering quality.}
  \label{fig:firstshow}
\end{figure}

To address these problems, we present Grid4D, a novel model with high dynamic scene rendering quality based on Gaussian splatting~\cite{gaussian-splatting}.
Our approach leverages hash encoding~\cite{instant-ngp} and proposes a new explicit representation method.
Unlike the plane-based explicit representations relying on the unsuitable low-rank assumption, as shown in Figure~\ref{fig:firstshow}, we decompose the 4D encoding into one spatial 3D hash encoding and three temporal 3D hash encodings.
Figure~\ref{fig:encoder} illustrates our proposed 4D decomposed hash encoding reduces overlap arising from the over-decomposed plane-based methods, resulting in more discriminative features.
Notably, our encoder generates two types of features: spatial features, representing static information across the timeline, and temporal features, capturing dynamic information.
For aggregation, we design a novel attention mechanism, directional attention, which leverages spatial features to generate attention scores in a directional range.
This directional attention aligns with the observation that deformation consistency within each scene component often varies across different components, and the attention from the spatial features could better help the model fit such differences.
However, like other explicit representation models, Grid4D often lacks smoothness.
To address this issue, we propose a novel training strategy incorporating smooth regularization which mitigates chaotic deformation predictions to enhance rendering clarity.

We compare Grid4D with several state-of-the-art dynamic scene rendering models.
Figure~\ref{fig:firstshow} and the experimental results show that Grid4D outperforms other models significantly in both visual quality and rendering speed.
In general, the contribution of this paper can be summarized as the following.
\begin{itemize}
    \item We propose a novel explicit representation method for dynamic scene rendering. By decomposing the 4D encoding into four 3D encodings, our 4D decomposed hash encoder effectively represents the features without relying on the low-rank assumption.
    \item We design a novel attention module for spatial and temporal feature aggregation. The directional attention module aligns with the variations in deformation consistency across different scene components, thereby enhancing deformation prediction accuracy.
    \item We employ a smooth training strategy to ensure the smoothness of our model. The smooth regularization effectively mitigates chaotic deformation predictions, resulting in high clarity in the rendered images produced by Grid4D.
\end{itemize}

\section{Related Works}\label{sec:related-works}

\textbf{NeRF-Based Dynamic Scene Rendering.}
NeRF~\cite{nerf} reconstructs light fields of static scenes by implicit representations and achieves significant visual improvements.
To extend NeRF capabilities for reconstructing dynamic scenes, applying implicit deformation fields to static models finds widespread use in dynamic scene rendering~\cite{d-nerf}.
To model dynamic scenes more accurately, various studies segment a scene into components with different attributes for different modeling~\cite{dynamic-nerf, nr-nerf}.
Moreover, several works apply higher-dimensional latent codes for the network input~\cite{dynerf, hypernerf} and incorporate additional supervision such as flow supervision across frames~\cite{fsdnerf, forwardflowdnerf, saff, nerflow, li2021neural, banmo, dynibar, 4dregsdf} and motion mask supervision~\cite{nerf-ds}. 
Meanwhile, focusing on modeling rigid objects is important in improving accuracy because of their unique physical properties and prevalence in most scenes~\cite{nr-nerf, star}.
Additionally, some research addresses the problems of dynamic scene models in several challenging scenes such as dynamic human modeling~\cite{ndr}, specular objects~\cite{nerf-ds} and the scenes without camera poses~\cite{rodynrf}.
However, implicit representations based on full MLPs suffer from the over-smoothing inherent property and require time-consuming training processes.
On the other hand, explicit representations, such as Triplanes~\cite{eg3d} and Hash Encoding~\cite{instant-ngp}, enhance NeRF by improving both visual quality and training speed.
A popular technique for plane-based explicit representations in dynamic scene rendering is decomposing 4D inputs into six 2D inputs~\cite{k-planes, hexplane, tensor4d, 4k4d, factorized-motion}. 
Also, hash encoding and 3D grid explicit representations can assist MLPs in predicting deformations with faster speed and higher precision~\cite{tineuvox, masked-spacetime-hashing, forwardflowdnerf, mixvoxels, park2023temporal, nerfplayer}.

\textbf{Gaussian-Based Dynamic Scene Rendering.}
Recently, Gaussian splatting~\cite{gaussian-splatting} models static scenes by Gaussian points, achieving both fast training and high visual quality.
When it comes to dynamic scene rendering, using 4D Gaussians or deforming Gaussians with pre-defined functions perform well in the cases with sufficient viewpoints~\cite{fudan4dgs, beida4dgs, gaussian-flow, spacetime-gaussians, dynamic-3d-gaussians}.
Alternatively, deforming the attributes of 3D Gaussians according to timestamps with neural networks has led to better outcomes in general dynamic scene rendering~\cite{deformable-3d-gaussians, 4d-gs, dynmf, md-splatting, gaufre, gags, sc-gs, cogs, 3dgstream}.
Fully MLP-based deformation fields achieve high rendering quality~\cite{deformable-3d-gaussians} but suffer from the over-smooth inherent property, resulting in the failure of some detail rendering and complex scenes.
Explicit representation models, for example, 4D-GS~\cite{4d-gs}, utilize the planes-based methods as the deformation field.
Although plane-based representations are more flexible, they are based on the unsuitable low-rank assumption, leading to massive feature overlap and rendering artifacts.
Our work mainly focuses on tackling the unsuitable low-rank assumption inherent in plane-based explicit representations to improve the rendering quality of Gaussian-based models.

\section{Method}\label{sec:method}


\subsection{Prelimaries: Gaussian Splatting}\label{sec:method-prelimaries}
Gaussian splatting~\cite{gaussian-splatting} is a static scene rendering model, known for its high training speed and visual quality.
This model assumes that the scene is composed of 3D Gaussian kernels with $\{\boldsymbol{\mu}, S, R, \sigma, \mathbf{c}\}$, corresponding to the position, scaling, rotation, opacity, and color.
Notably, the color attribute is defined by the spherical harmonic coefficients~(SH).
To render the scene, by using a view transform matrix $W$ and a projective Jacobian matrix $J$, Gaussians can be splatted onto camera planes~\cite{splatting, differentiable-splatting}.
\begin{align}
\Sigma' = JW\Sigma W^TJ^T,\ \Sigma = RSS^TR^T
\end{align}
where $\Sigma'$ is the covariance matrix in camera planes and $\Sigma$ is the original Gaussian covariance which can be calculated by the scaling and rotation attributes.
Finally, supposing that the pixel on the camera planes is $\mathbf{p}$, the splatted Gaussians can be rendered by the volume rendering equation,
\begin{align}
\mathbf{C}(\mathbf{p}) = \sum_{i=1}^N \mathbf{c}_i \alpha_i \prod_{j=1}^{i-1}(1-\alpha_j), \alpha_i = \sigma_i e^{-\frac{1}{2}(\mathbf{p}-\boldsymbol{\mu}^p_i)^T\Sigma'^{-1}_i(\mathbf{p}-\boldsymbol{\mu}^p_i)}
\end{align}
where $\boldsymbol{\mu}^p$ is the projected coordinates of the 3D Gaussians, and $N$ is the number of overlapped Gaussians on the pixel.

In optimization, adaptive density control is crucial for convergence.
It involves pruning low-opacity Gaussians and densifying them based on the gradients and scaling.
However, original Gaussian splatting cannot represent dynamic scenes and needs the help of deformation fields.

\subsection{4D Decomposed Hash Encoding}\label{sec:method-4d-decomposed-hash}

\begin{figure}[tb]
  \centering
  \includegraphics[scale=0.62, trim=0 355 325 0, clip]{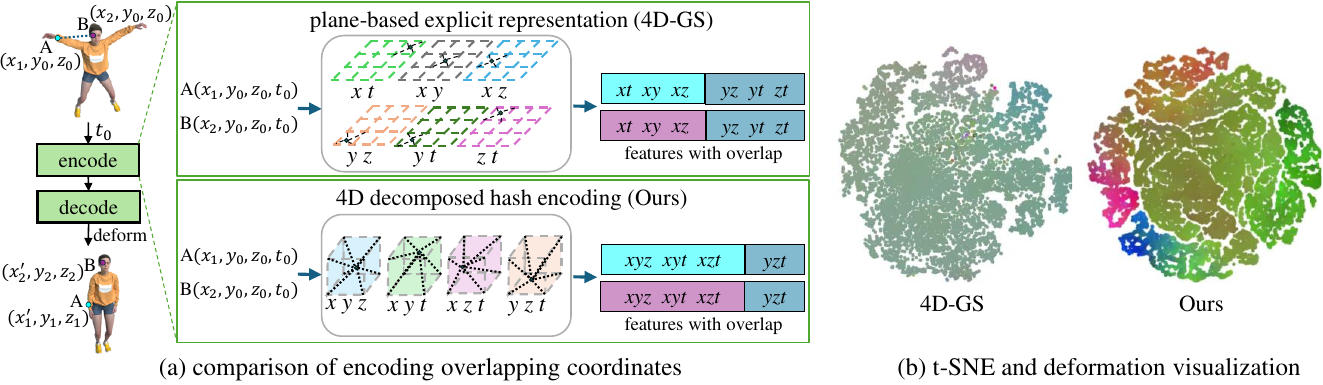}
  \caption{Comparison of our proposed 4D decomposed hash encoding with the plane-based explicit representation~\cite{4d-gs}. (a) Compared to the plane-based methods based on the low-rank assumption, our methods reduce the overlap ratio in the features from a half to a quarter when encoding points A and B with heavily overlapping coordinates. (b) is the t-SNE~\cite{t-sne} visualization of all the features, and the colors denote the corresponding represented deformations. The diversity of colors demonstrates that the reduced overlap makes the features represent different deformations more effectively.
  }
  \label{fig:encoder}
\end{figure}

Dynamic scene rendering involves deforming Gaussians according to a 4D coordinate $(x, y, z, t)$ input, where $t$ represents the timestamp and $(x, y, z)$ means the position of a Gaussian.
Instead of employing the over-smooth fully MLP-based implicit representations, we use explicit representation for Grid4D.
However, existing plane-based explicit representation relies on the unsuitable low-rank assumption which overly decomposes the $(x, y, z, t)$ encoding into $(x, y), (y, z), (x, z), (x, t), (y, t), (z, t)$ plane encodings~\cite{tensor4d, k-planes, hexplane, 4d-gs}.
As shown in Figure~\ref{fig:encoder}(a), for instance, considering the Gaussians A and B with the same $y$ and $z$ coordinates.
The plane-based method has the same encoded features in the $(y, z), (y, t), (z, t)$ planes.
Such a high overlap ratio might lead to the low discriminability of the features and block the model from fitting different deformations accurately.
To address this problem, directly removing the decomposition by simply adding the time dimension to the traditional 3D grid for the 4D hyper-grid hash encoding is a possible way.
However, the 4D hyper-grid hash encoding leads to high collision rates due to the high space complexity $O(n^4)$ of the 4D hyper-grid~\cite{masked-spacetime-hashing}.
Therefore, thoroughly eliminating the overlap might not be an available solution.

\textbf{Tri-axial 4D Decomposed Grid.}
To address this problem, we propose a novel decomposition approach that decomposes the 4D encoding $(x, y, z, t)$ into four 3D hash encodings $(x, y, z), (x, y, t), (y, z, t), (x, z, t)$.
The decomposition allows us to work with fewer parameters, which reduces the space complexity from $O(n^4)$ to $O(n^3)$ without relying on the low-rank assumption.
As shown in Figure~\ref{fig:encoder}(a), the tri-axial decomposition can effectively reduce the overlap ratio from a half to a quarter, thereby enhancing each feature to represent the corresponding deformation.
Figure~\ref{fig:encoder}(b) demonstrates that the features encoded by our methods are more discriminative for deformation prediction than plane-based methods.

\textbf{Multiresolution Hash Encoding.}
In the original hash encoding technique~\cite{instant-ngp}, the grid employed in the encoder has the same resolution across all dimensions.
Consistent resolutions could be suitable for static scene rendering, where the isotropic sampling assumption holds in the 3D space.
Nevertheless, the sampling of the 4D space is usually anisotropic, which is usually sparse in the time dimension.
Therefore, in our implementation, the temporal 3D encodings~$(x, y, t), (y, z, t), (x, z, t)$ have different resolutions in the $t$ dimension to account for this sparsity.
Following the InstantNGP~\cite{instant-ngp}, we set the multiple resolutions of each dimension in a geometric progression:
\begin{align}
    N_l = \lfloor N_{min} \cdot b \rfloor, \ b = \exp{\left(\frac{\ln{N_{max}} - \ln{N_{min}}}{L - 1}\right)}
\end{align}
where $N_{min}, N_{max}$ is the coarsest and finest resolutions, $l$ is level number, $L$ is the max level, and $N_l$ is the resolution we select.
The grid voxel positions for the input $\mathbf{x}$ could be calculated by rounding down and up in each level $\lfloor \mathbf{x}_l \rfloor = \lfloor \mathbf{x} \cdot N_l \rfloor, \lceil \mathbf{x}_l \rceil = \lceil \mathbf{x} \cdot N_l \rceil$.
The voxels in each level could be obtained from the hash table by hashing the corresponding positions:
\begin{align}
h_l(\mathbf{x}_l) = \left( \bigoplus_{i=1, x_i \in \mathbf{x}_l}^d x_i \pi_i \right) \mod T_l
\end{align}
where $\bigoplus$ is the bit-wise XOR operation, $d$ is the dimension of the input, $\pi_i$ are unique large prime numbers, $T_l$ is the size of the level $l$ hash table.
Then the encoded features could be calculated by the trilinear interpolation of the grid voxel values.
Generally, the encoded features of the 4D input $(x, y, z, t)$ include the spatial and temporal features from the spatial grid hash encoder~$G_{xyz}$ and temporal grid hash encoders~$G_{xyt}, G_{yzt}, G_{xzt}$ respectively.

\subsection{Multi-head Directional Attention Decoder}\label{sec:method-directional-attention}

\begin{figure}
  \centering
  \includegraphics[scale=0.49, trim=60 180 95 20, clip]{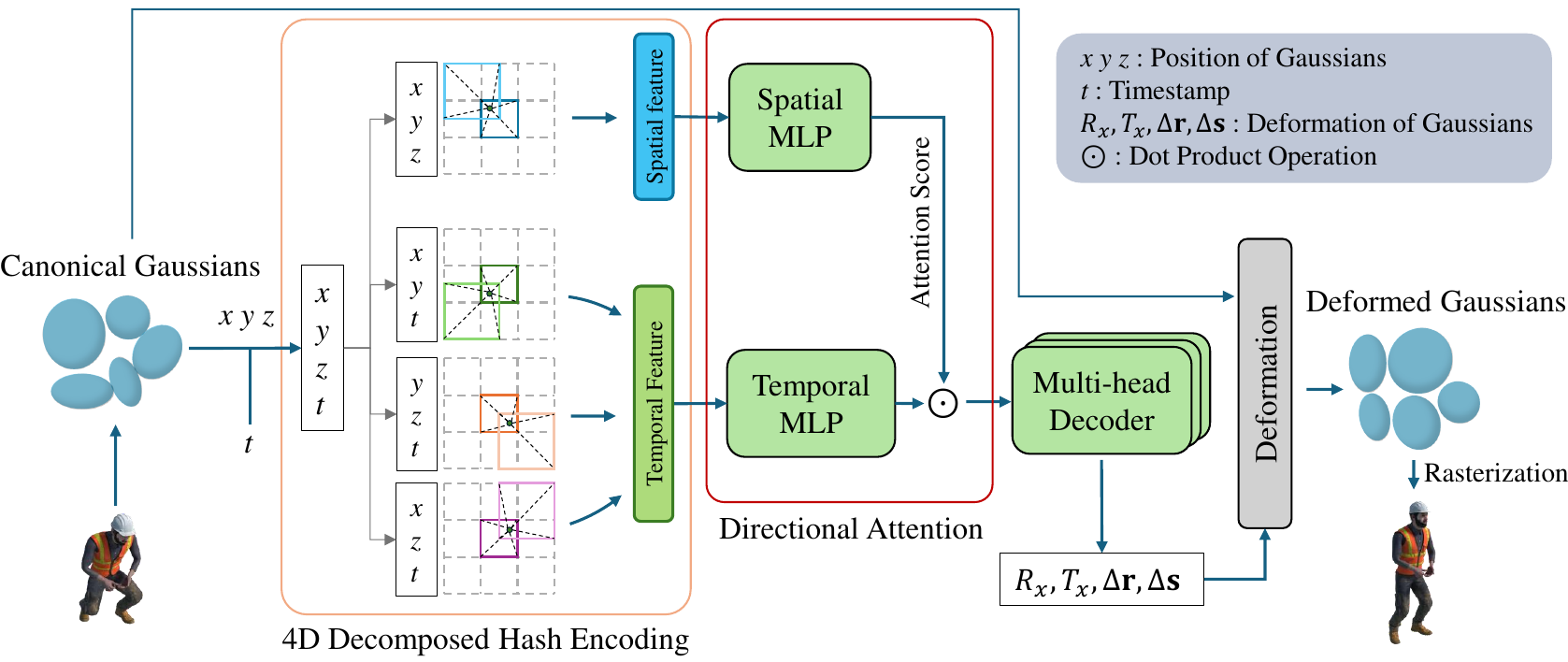}
  \caption{The overview of Grid4D. Given the canonical Gaussians and the timestamp, we first encode the decomposed input separately. Then we apply the directional attention scores generated by the spatial static features to the temporal dynamic features, and we decode the features with a tiny multi-head MLP. 
  Finally, the Gaussians deformed by the predicted deformations are splatted by the differentiable rasterization operation~\cite{gaussian-splatting} to render the images for supervision.}
  \label{fig:pipeline}
\end{figure}

Our 4D decomposed hash encoding generates two types of features: temporal features and spatial features.
The temporal features represent the information related to the timestamp while the spatial features represent the common information across the timeline.
The Gaussians representing different scene components often have various deformations in almost every timestamp.
Therefore, the spatial features could be used to help the model fit such variations, and we design the directional attention module for the spatial and temporal feature aggregation.

\textbf{Directional Attention.}
We infer the attention features from the spatial grid hash encoder~$G_{xyz}$ with a tiny spatial MLP~$f_{s}$, and generate the score $\mathbf{a}$ through the following formula,
\begin{align}
\mathbf{a} = 2\cdot \Phi(\mathbf{h}_{xyz}) - 1,\ \mathbf{h}_{xyz} = f_{s}\circ G_{xyz}(x, y, z)
\end{align}
where $\Phi$ is the Sigmoid function.
We consider that several components probably have entirely opposite deformations against the neighboring Gaussians.
For example, the Gaussians for the shadows often have opposite motions relative to the objects.
Therefore, different from the common range $(0, 1)$ of the attention score~$\mathbf{a}$, we scale it to a directional range~$(-1, 1)$ to represent neighboring deformations with opposite directions, thereby enhancing the representation ability of the attention mechanism.

Then we apply the attention score to the activated deformation features encoded by the three temporal grid hash encoders~$G_{xyt}, G_{yzt}, G_{xzt}$ and a tiny temporal MLP~$f_{t}$.
\begin{align}
\mathbf{h} = \mathbf{a} \odot f_t(G_{xyt}(x, y, t), G_{yzt}(y, z, t), G_{xzt}(x, z, t))
\end{align}
where $\odot$ is the dot product operation.
Finally, we get the deformation features~$\mathbf{h}$ with high representation ability.
Our experiments demonstrate that our attention module outperforms the architecture which either directly decodes the concatenation of the spatial and temporal features or uses the common range $(0, 1)$ of the attention score.

\textbf{Multi-head Deformation Decoder.}
The decoder is required to decode the features~$\mathbf{h}$ to get the Gaussian deformation.
Different from the prior works~\cite{4d-gs, deformable-3d-gaussians}, we use a tiny multi-head MLP~$D$ to decode the features and predict the position deformation with a rotation matrix~$R_x$ and a translation matrix~$T_x$ as~\cite{sc-gs}.
Finally, we deform the position, scaling, and rotation of the Gaussians in the canonical space with the predicted deformation.
\begin{align}
\boldsymbol{\mu}' = R_x\boldsymbol{\mu} + T_x, \ S' = S + \Delta \mathbf{s}, \ R' = R + \Delta \mathbf{r}, \ D(\mathbf{h}) = \{R_x, T_x, \Delta \mathbf{r}, \Delta \mathbf{s}\},
\end{align}
We define the rotation matrix~$R_x$ with a quaternion for more accurate interpolation and stable optimization.
Following Gaussian splatting~\cite{gaussian-splatting}, the deformed Gaussians could be rendered into images of specific timestamps via differentiable rasterization.

\subsection{Training with Smooth Regularization}\label{sec:method-smooth-regularization}

Although the proposed model architecture could effectively predict the Gaussian deformation, the 4D decomposed hash encoder still suffers from the lack of smoothness, a common challenge in most explicit representation methods.
We consider that the MLP decoder has the smooth inherent property and does not require additional smoothing.
Therefore, we set our regularization in the feature space without involving the MLP decoder inference for higher efficiency.
Generally, to regularize the hash encoder, we propose a novel smooth regularization loss.
\begin{align}\label{equa:smooth}
    \mathcal{L}_r = ||G_{xyzt}(x, y, z, t) - G_{xyzt}(x + \epsilon_x, y + \epsilon_y, z + \epsilon_z, t + \epsilon_t)||_2^2
\end{align}
where ($\epsilon_x, \epsilon_y, \epsilon_z, \epsilon_t$) is the small random perturbation for the input $(x, y, z, t)$ respectively, and $G_{xyzt}$ is the concatenation of four grid hash encoders.
This regularization enforces similarity among encoded features in neighboring regions, thereby making the nearby Gaussians have similar deformations.
Due to the difference of spatial and temporal encoding, we use a different regularization setting for the spatial encoding for several cases.
Notably, to improve the efficiency, we randomly select partial Gaussians for the regularization instead of using them all.
Our experiments demonstrate that this smooth regularization effectively mitigates the deformation chaos, leading to significantly improved rendering clarity.

In general, similar to Gaussian splatting~\cite{gaussian-splatting}, our total loss function can be summarized as the weighted sum of L1 color loss, D-SSIM loss, and the proposed smooth regularization term.
\begin{align}\label{equa:total-loss}
\mathcal{L} = (1 - \lambda_c) \mathcal{L}_1 + \lambda_c \mathcal{L}_{D-SSIM} + \lambda_r \mathcal{L}_r
\end{align}
where $\lambda_c, \lambda_r$ are the hyperparameters to balance the losses.
Following~\cite{deformable-3d-gaussians}, we use the detached Gaussian positions for deformation prediction, which results in better performance.
Also, similar to prior works~\cite{deformable-3d-gaussians, 4d-gs}, we initialize the static canonical Gaussians without deformation at the beginning of the training process.
Specifically for SfM~\cite{colmap} initialized Gaussians, we shorten or remove the static initialization process.
We apply the same adaptive density controller and opacity resetting mechanism as Gaussian splatting~\cite{gaussian-splatting}.
The pipeline of Grid4D is illustrated by Figure~\ref{fig:pipeline}.

\section{Experiments}\label{sec:experiments}

In this section, we introduce our experiments conducted on a single RTX 3090 GPU.
We build our code mainly on PyTorch~\cite{pytorch}, while we implement our 4D decomposed hash encoder with CUDA/C++.
More experimental results and analysis can be found in the supplementary.

\subsection{Experimental Setup}\label{sec:experiments-setup}

\begin{figure}[tb]
  \centering
  \includegraphics[scale=0.65, trim=55 420 25 10, clip]{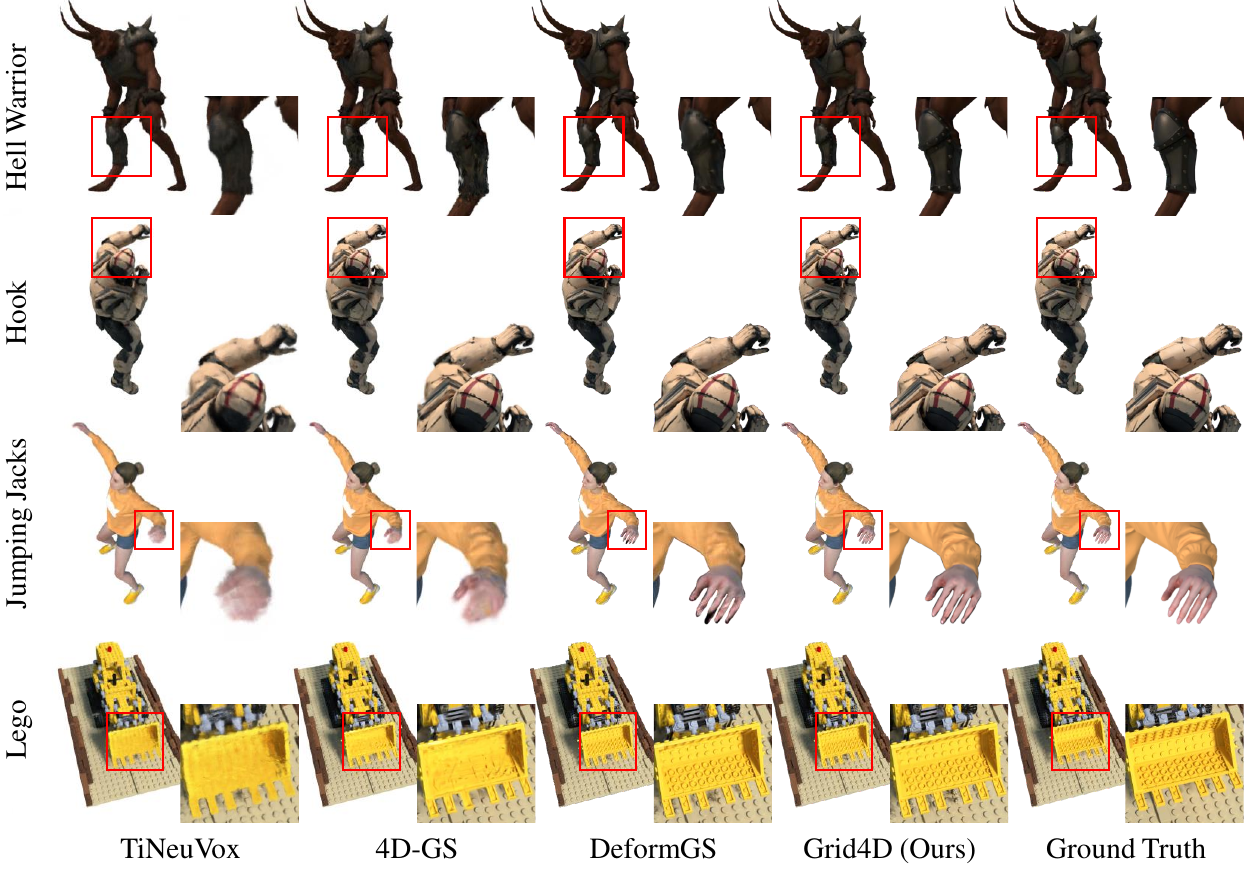}
  \caption{Qualitative comparisons on the synthetic D-NeRF dataset~\cite{d-nerf} with our baselines~\cite{tineuvox, 4d-gs, deformable-3d-gaussians}.}
  \label{fig:dnerf}
\end{figure}

\textbf{Datasets.}
We evaluate Grid4D on two popular datasets.
D-NeRF~\cite{d-nerf} dataset is a public monocular synthetic dataset that provides accurate and time-varying camera poses.
HyperNeRF~\cite{hypernerf} dataset is a public real-world dataset captured by one or two moving cameras.
Neu3D~\cite{dynerf} dataset is a public dataset captured by multiple cameras with fixed poses.
However, different from synthetic datasets, the camera poses of the HyperNeRF and Neu3D datasets are estimated by COLMAP~\cite{colmap}, which is not accurate.
We set the rendering resolutions of the D-NeRF, HyperNeRF and Neu3D datasets to~$800\times 800$, $536\times900$ and $1352\times1024$ respectively.
Notably, we find several mistakes in the ground truth of the `Lego' scene in the D-NeRF dataset, as shown in the last row of Figure~\ref{fig:dnerf}, so we ignore this scene in all quantitative comparisons of rendering quality.

\textbf{Baselines.}
We compare Grid4D with several state-of-the-art models~\cite{hexplane, tineuvox, 4d-gs, deformable-3d-gaussians, sc-gs}.
HexPlane~\cite{hexplane} and TiNeuVox~\cite{tineuvox} are NeRF-based dynamic scene rendering models, utilizing plane-based and 3D grid explicit representations respectively. 
4D-GS~\cite{4d-gs} and DeformGS~\cite{deformable-3d-gaussians} are Gaussian-based models, employing plane-based explicit representation and fully MLP-based implicit representation for the deformation fields respectively.
SC-GS~\cite{sc-gs} is a model built on DeformGS~\cite{deformable-3d-gaussians}, and proposes to use sparse control points for better dynamic scene rendering and edit.

\textbf{Hyperparameters.}
For all datasets, we configure the resolution of the spatial grid hash encoder to span from $16$ to $2048$ across $16$ levels.
Meanwhile, the max level number $L$ of temporal grid hash encoders remains consistent at $32$.
We set $\lambda_c$ and $\lambda_r$ to $0.2$ and $0.5$ for common scenes and follow a similar learning rate schedule as DeformGS~\cite{deformable-3d-gaussians, gaussian-splatting}.

\begin{table}[tb]
\caption{Quantitative comparisons on the synthetic D-NeRF~\cite{d-nerf} dataset. The higher PSNR~$(\uparrow)$, higher SSIM~$(\uparrow)$ and lower LPIPS~$(\downarrow)$ denote better rendering quality. The color of each cell shows the \sethlcolor{pink!150}\hl{best} and the \sethlcolor{orange!40}\hl{second best}.}
\label{tab:comparison-d-nerf}
\centering
\small
\setlength{\tabcolsep}{0.9mm}{
\begin{tabular}{ccccccccccccc}
\toprule
         & \multicolumn{3}{c}{Bouncing Balls} & \multicolumn{3}{c}{Hell Warrior} & \multicolumn{3}{c}{Hook} & \multicolumn{3}{c}{Jumping Jacks} \\
Model                    &  PSNR &  SSIM  &  LPIPS &  PSNR  &  SSIM  & LPIPS  & PSNR & SSIM & LPIPS & PSNR & SSIM & LPIPS \\
\midrule
HexPlane~\cite{hexplane} & 40.36 & 0.992 & 0.031 & 24.30 & 0.944 & 0.073 & 28.26 & 0.955 & 0.052 & 31.74 & 0.974 & 0.036 \\
TiNeuVox~\cite{tineuvox} & 40.28 & 0.992 & 0.042 & 27.29 & 0.964 & 0.076 & 30.51 & 0.959 & 0.060 & 33.46 & 0.977 & 0.041 \\
4D-GS~\cite{4d-gs} & 40.77 & 0.994 & 0.015 & 28.80 & 0.974 & 0.037 & 32.95 & 0.977 & 0.027 & 35.50 & 0.986 & 0.020 \\
DeformGS~\cite{deformable-3d-gaussians} & 40.91 & \cellcolor{orange!40} 0.995 & \cellcolor{orange!40} 0.009 & 41.34 & 0.987 & 0.024 & 37.06 & 0.986 & 0.016 & 37.66 & 0.989 & \cellcolor{orange!40} 0.013 \\
SC-GS~\cite{sc-gs} & 41.59 & \cellcolor{orange!40} 0.995 & \cellcolor{orange!40} 0.009 & 42.19 & 0.989 & 0.019 & 38.79 & \cellcolor{orange!40} 0.990 & 0.011 & 39.34 & \cellcolor{orange!40} 0.992 & \cellcolor{pink!150} 0.008 \\
Grid4D~(Ours) & \cellcolor{pink!150} 42.62 & \cellcolor{pink!150} 0.996 & \cellcolor{pink!150} 0.008 & \cellcolor{pink!150} 42.85 & \cellcolor{pink!150} 0.991 & \cellcolor{pink!150} 0.015 & \cellcolor{orange!40} 38.89 & \cellcolor{orange!40} 0.990 & \cellcolor{orange!40} 0.009 & \cellcolor{orange!40} 39.37 & \cellcolor{pink!150} 0.993 & \cellcolor{pink!150} 0.008 \\
Grid4D + SC & \cellcolor{orange!40} 42.17 & \cellcolor{orange!40} 0.995 & \cellcolor{pink!150} 0.008 & \cellcolor{orange!40} 42.81 & \cellcolor{orange!40} 0.990 & \cellcolor{orange!40} 0.017 & \cellcolor{pink!150} 40.26 & \cellcolor{pink!150} 0.992 & \cellcolor{pink!150} 0.008 & \cellcolor{pink!150} 39.58 & \cellcolor{pink!150} 0.993 & \cellcolor{pink!150} 0.008 \\
\toprule
         & \multicolumn{3}{c}{Mutant} & \multicolumn{3}{c}{Standup} & \multicolumn{3}{c}{Trex} & \multicolumn{3}{c}{Mean}\\
Model    &  PSNR  &  SSIM  & LPIPS  & PSNR & SSIM & LPIPS & PSNR & SSIM & LPIPS & PSNR & SSIM & LPIPS \\
\midrule
HexPlane~\cite{hexplane} & 33.66 & 0.982 & 0.028 & 34.12 & 0.983 & 0.019 & 31.01 & 0.976 & 0.028 & 31.92 & 0.972 & 0.038 \\
TiNeuVox~\cite{tineuvox} & 32.07 & 0.961 & 0.048 & 34.46 & 0.980 & 0.033 & 31.43 & 0.967 & 0.047 & 32.78 & 0.972 & 0.050 \\
4D-GS~\cite{4d-gs} & 37.75 & 0.988 & 0.016 & 38.15 & 0.990 & 0.014 & 33.95 & 0.985 & 0.022 & 35.41 & 0.985 & 0.021 \\
DeformGS~\cite{deformable-3d-gaussians} & 42.47 & \cellcolor{orange!40} 0.995 & \cellcolor{orange!40} 0.005 & 44.14 & \cellcolor{orange!40} 0.995 & \cellcolor{orange!40} 0.007 & 37.56 & 0.993 & 0.010 & 40.16 & 0.991 & 0.012 \\
SC-GS~\cite{sc-gs} & 43.43 & \cellcolor{pink!150} 0.996 & \cellcolor{orange!40} 0.005 & \cellcolor{orange!40} 46.72 & \cellcolor{pink!150} 0.997 & \cellcolor{pink!150} 0.004 & 39.53 & \cellcolor{orange!40} 0.994 & \cellcolor{orange!40} 0.009 & 41.65 & \cellcolor{orange!40} 0.993 & \cellcolor{orange!40} 0.009 \\
Grid4D~(Ours) & \cellcolor{orange!40} 43.94 & \cellcolor{pink!150} 0.996 & \cellcolor{pink!150} 0.004 & 46.28 & \cellcolor{pink!150} 0.997 & \cellcolor{pink!150} 0.004 & \cellcolor{orange!40} 40.01 & \cellcolor{pink!150} 0.995 & \cellcolor{pink!150} 0.008 & \cellcolor{orange!40} 42.00 & \cellcolor{pink!150} 0.994 & \cellcolor{pink!150} 0.008 \\
Grid4D + SC & \cellcolor{pink!150} 44.07 & \cellcolor{pink!150} 0.996 & \cellcolor{pink!150} 0.004 & \cellcolor{pink!150} 46.87 & \cellcolor{pink!150} 0.997 & \cellcolor{pink!150} 0.004 & \cellcolor{pink!150} 41.12 & \cellcolor{pink!150} 0.995 & \cellcolor{pink!150} 0.008 & \cellcolor{pink!150} 42.41 & \cellcolor{pink!150} 0.994 & \cellcolor{pink!150} 0.008 \\
\bottomrule
\end{tabular}}
\end{table}

\subsection{Comparisons}\label{sec:experiments-comparisons}

\begin{table}[!tb]
\caption{Quantitative comparison on the validation rig part~(Rig) and the interpolation part~(Interpolation) of the real-world HyperNeRF~\cite{hypernerf} dataset. The higher PSNR~$(\uparrow)$ and higher MS-SSIM~$(\uparrow)$ denote better rendering quality. The color of each cell shows the \sethlcolor{pink!150}\hl{best} and the \sethlcolor{orange!40}\hl{second best}.}
\label{tab:comparison-hypernerf}
\centering
\small
\begin{tabular}{ccccc}
\toprule
         & \multicolumn{2}{c}{Rig(4 scenes)} & \multicolumn{2}{c}{Interpolation(6 scenes)} \\
Model                    &  PSNR &  MS-SSIM  &  PSNR  &  MS-SSIM \\
\midrule
TiNeuVox~\cite{tineuvox} &  24.20 & 0.836 & 27.08 & \cellcolor{orange!40} 0.922 \\
4D-GS~\cite{4d-gs} & \cellcolor{orange!40} 24.99 & \cellcolor{orange!40} 0.838 & \cellcolor{orange!40} 27.54 & 0.912 \\
Grid4D~(Ours) & \cellcolor{pink!150} 25.50 & \cellcolor{pink!150} 0.856 & \cellcolor{pink!150} 28.56 & \cellcolor{pink!150} 0.933 \\
\bottomrule
\end{tabular}
\end{table}

\begin{figure}[!tb]
    \centering
    \includegraphics[scale=0.55, trim=0 255 250 0, clip]{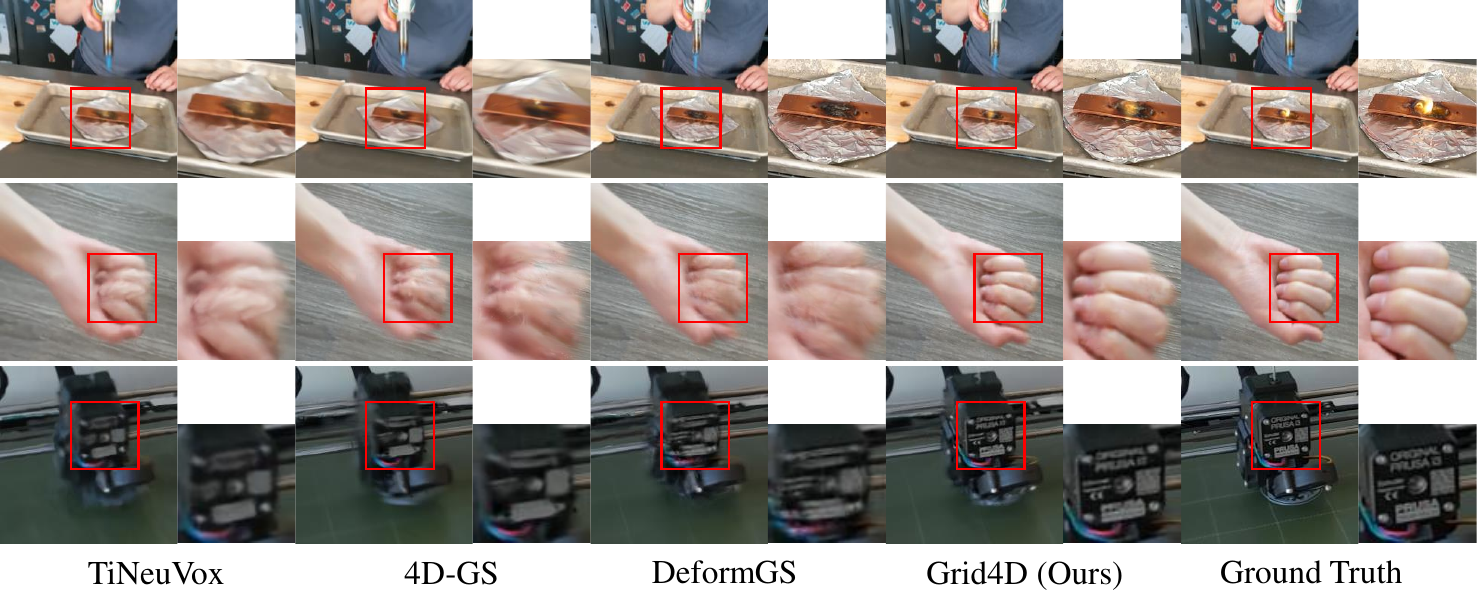}
    \caption{Qualitative comparisons on the real-world HyperNeRF~\cite{hypernerf} dataset.}
    \label{fig:hypernerf}
\end{figure}

\begin{table}[!tb]
\caption{Quantitative comparison on the real-world Neu3D~\cite{dynerf} dataset. The higher PSNR~$(\uparrow)$ and higher SSIM~$(\uparrow)$ denote better rendering quality. The color of each cell shows the \sethlcolor{pink!150}\hl{best}.}
\label{tab:comparison-neu3d}
\centering
\small
\setlength{\tabcolsep}{0.9mm}{
\begin{tabular}{ccccccccccccc}
\toprule
         & \multicolumn{2}{c}{Coffee Martini} & \multicolumn{2}{c}{Cook Spinach} & \multicolumn{2}{c}{Cut Beef} & \multicolumn{2}{c}{Flame Salmon} & \multicolumn{2}{c}{Flame Steak} & \multicolumn{2}{c}{Sear Steak} \\
    Model    & PSNR & SSIM & PSNR & SSIM & PSNR & SSIM & PSNR & SSIM & PSNR & SSIM & PSNR & SSIM \\
\midrule
4D-GS~\cite{4d-gs} & 27.34 & \cellcolor{pink!150} 0.898 & 32.50 & 0.942 & 32.26 & 0.942 & 27.99 & 0.902 & 32.54 & 0.951 & \cellcolor{pink!150} 33.44 & 0.954 \\
Grid4D~(Ours) & \cellcolor{pink!150} 28.30 & \cellcolor{pink!150} 0.898 & \cellcolor{pink!150} 32.58 & \cellcolor{pink!150} 0.948 & \cellcolor{pink!150} 33.22 & \cellcolor{pink!150} 0.950 & \cellcolor{pink!150} 29.12 & \cellcolor{pink!150} 0.908 & \cellcolor{pink!150} 32.56 & \cellcolor{pink!150} 0.955 & 33.16 & \cellcolor{pink!150} 0.957 \\
\bottomrule
\end{tabular}}
\end{table}

\begin{figure}[!tb]
    \centering
    \includegraphics[scale=0.46, trim=0 235 100 0, clip]{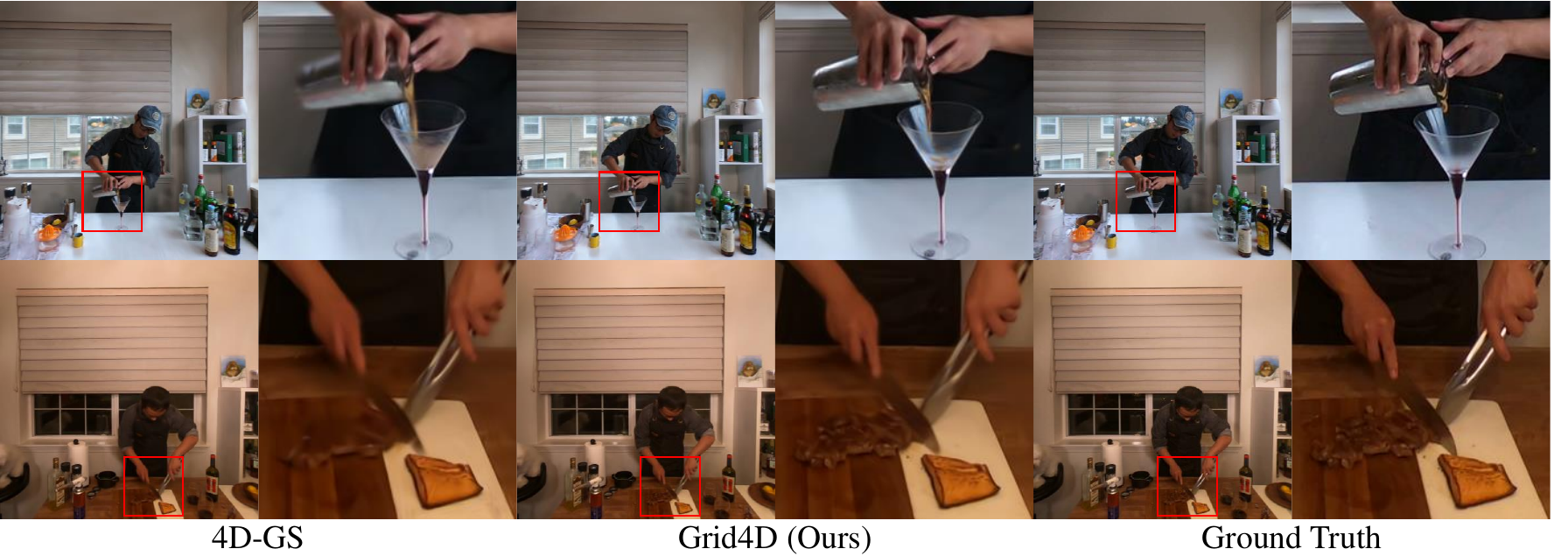}
    \caption{Qualitative comparisons on the real-world Neu3D~\cite{dynerf} dataset.}
    \label{fig:neu3d}
\end{figure}

\begin{table}[!tb]
\caption{Rendering speed comparison on the synthetic D-NeRF~\cite{d-nerf} dataset. We report the FPS based on the number of Gaussian points. Compared to other models, our model still achieves high rendering speed and real-time rendering when facing a much larger amount of Gaussians.}
\label{tab:comparison-speed}
\centering
\small
\setlength{\tabcolsep}{1.8mm}{
\begin{tabular}{ccccccccc}
\toprule
FPS / Num(k) & Balls & Warrior & Hook & Jumping & Lego & Mutant & Standup & Trex \\
\midrule
4D-GS~\cite{4d-gs} & 182 / 28 & 168 / 40 & 91 / 39 & 207 / 24  & 104 / 93 & 173 / 38 & 201 / 27 & 151 / 68 \\
DeformGS~\cite{deformable-3d-gaussians} & 37 / 180 & 161 / 37 & 43 / 150 & 71 / 90 & 30 / 289 & 49 / 169 & 77 / 81 & 30 / 217 \\
Grid4D~(Ours) & 91 / 192 & 334 / 46 & 79 / 210 & 241 / 68 & 64 / 302 & 157 / 126 & 170 / 100 & 86 / 254 \\
\bottomrule
\end{tabular}}
\end{table}

\textbf{Comparison of Visual Quality.}
We compare Grid4D with the state-of-the-art models on the synthetic D-NeRF~\cite{d-nerf} dataset~(Table~\ref{tab:comparison-d-nerf} and Figure~\ref{fig:dnerf}), the real-world HyperNeRF~\cite{hypernerf} dataset~(Table~\ref{tab:comparison-hypernerf} and Figure~\ref{fig:hypernerf}) and the real-world Neu3D~\cite{dynerf} dataset~(Table~\ref{tab:comparison-neu3d} and Figure~\ref{fig:neu3d}).
The PSNR, SSIM~\cite{ssim}, LPIPS~\cite{lpips}(VGG~\cite{vgg}), and MS-SSIM are the metrics denoting visual quality.
Notably, the DeformGS~\cite{deformable-3d-gaussians} model fails to construct several HyperNeRF scenes with large motions and imprecise camera poses, as mentioned in their paper.
Several failed cases can be found in Section~\ref{sec:supp-comparison} of the supplementary, and we consider that this is also due to the over-smooth inherent property of fully MLP-based implicit representation.

Due to the inherent flexibility of the explicit representation, the results of the `Hook' scene show that Grid4D has a stronger ability to reconstruct fine structures than DeformGS~\cite{deformable-3d-gaussians} which is based on the implicit representation.
We also apply the sparse control points in SC-GS~\cite{sc-gs} to our model and build SC-GS on Grid4D rather than DeformGS for further evaluation. 
We refer to it as 'Grid4D + SC', and observe an improvement in comparison to Grid4D and SC-GS as list in the last three rows of Table~\ref{tab:comparison-d-nerf}.
Thanks to our 4D decomposed hash encoding, when facing the scenes with complex motions and Gaussians with heavily overlapping coordinates, such as `JumpingJacks', Grid4D predicts the deformations much more accurately than 4D-GS~\cite{4d-gs} which is built on the planed-based explicit representation relying on the unsuitable low-rank assumption.

\textbf{Comparison of Rendering Speed.}
Comparing Frames Per Second~(FPS) directly might not be a fair experiment because the number of Gaussians is quite different among different models.
Therefore, we list both the FPS and the corresponding Gaussian count in Table~\ref{tab:comparison-speed}.
Despite the acceleration provided by the CUDA/C++ implementation in Grid4D, our proposed explicit representation makes Grid4D exhibit significantly faster rendering performance.
When facing a huge number of Gaussians, Grid4D maintains high rendering speed and achieves real-time rendering.
However, Grid4D has no improvement in the training speed compared to DeformGS~\cite{deformable-3d-gaussians}.
Although we do not use all Gaussians for the regularization, the smooth regularization requires Grid4D to encode the input twice, which slows down the training process.
Meanwhile, we find that Grid4D’s accurate deformation predictions often lead to an increase in the number of Gaussians, contributing to time overhead.
Nevertheless, Grid4D needs less GPU memory for training than DeformGS~\cite{deformable-3d-gaussians}, and the extra computational cost has little influence on training in comparison to the significant improvements in rendering quality.
More details can be found in Section~\ref{sec:supp-limitation} of our supplementary.

\subsection{Ablation Study and Analysis}\label{sec:experiments-ablation}

\begin{figure}[tb]
  \centering
  \includegraphics[scale=0.6, trim=0 333 350 0, clip]{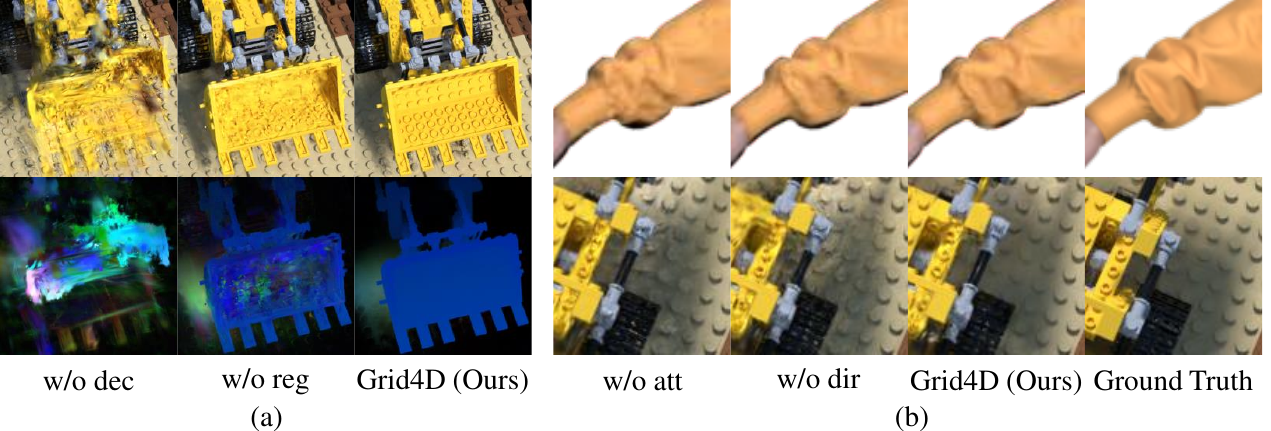}
  \caption{Qualitative results of our ablation studies. (a) is the ablation of the 4D decomposed hash encoding and the smooth regularization. The first row is the rendering results, and the second row is the visualization of the deformation. The similar colors in the deformation map mean similar deformation sizes on each axis. (b) is the ablation of the directional attention.}
  \label{fig:ablation}
\end{figure}

To mitigate the distraction of imprecise camera poses, we mainly conduct our ablation studies on the synthetic D-NeRF~\cite{d-nerf} dataset. 
Table~\ref{tab:ablation-d-nerf} and Figure~\ref{fig:ablation} show the results of our ablations.

\textbf{Ablation of 4D Decomposed Hash Encoding.}
The proposed 4D decomposed hash encoding splits the 4D input into four 3D inputs, encoding them separately without the unsuitable low-rank assumption and high space complexity.
To demonstrate the advantages of our encoding method, we employ the simple 4D hyper-grid hash encoding in Grid4D w/o dec.
The chaos in Figure~\ref{fig:ablation}(a) illustrates the rendering degradation caused by the high hash collision rate.

\begin{table}[tb]
\caption{Quantitative ablation results on the synthetic D-NeRF~\cite{d-nerf} dataset. The color of each cell shows the \sethlcolor{pink!150}\hl{best} and the \sethlcolor{orange!40}\hl{second best}.}
\label{tab:ablation-d-nerf}
\centering
\small
\begin{tabular}{ccccccc}
\toprule
Model                    & w/o dec & w/o reg & w/o att & w/o dir & Grid4D~(Ours) \\
\midrule
PSNR & 28.45 & 39.47 & \cellcolor{orange!40} 41.37 & 41.32 & \cellcolor{pink!150} 42.00 \\
SSIM & 0.949 & 0.991 & \cellcolor{orange!40} 0.993 & \cellcolor{orange!40} 0.993 & \cellcolor{pink!150} 0.994 \\
LPIPS & 0.055 & 0.012 & \cellcolor{orange!40} 0.009 & \cellcolor{orange!40} 0.009 & \cellcolor{pink!150} 0.008 \\
\bottomrule
\end{tabular}
\end{table}

\textbf{Ablation of Directional Attention.}
The directional attention helps Grid4D accurately predict the different deformations across different scene components.
We scale the attention score to~$(0, 1)$ for Grid4D w/o dir to demonstrate the advantage of the directional range~$(-1, 1)$.
We also compare our attention module with the simple architecture Grid4D w/o att which directly decodes the concatenation of the spatial and temporal features.
In Figure~\ref{fig:ablation}(b), the shadow has obvious different deformations from the neighboring parts across the timeline.
Our directional attention achieves high clarity in rendering the portion with the shadow, emphasizing its effectiveness in capturing such variations.


\textbf{Ablation of Smooth Regularization.}
The proposed smooth regularization aims at mitigating the chaos of deformation prediction.
We train Grid4D without the smooth regularization in Equation~\ref{equa:smooth} and refer to the model as Grid4D w/o reg.
The results in Figure~\ref{fig:ablation}(a) show that the regularization reduces the deformation artifacts caused by the lack of smoothness.

We conduct more ablation studies for the architecture and smooth regularization.
We also visualize the intermediate results of our model.
More results can be found in Section~\ref{sec:supp-ablation} of our supplementary.






\section{Conclusion}\label{sec:conclusion}
In this paper, we have introduced Grid4D, a novel model for high-fidelity dynamic scene rendering.
Grid4D utilizes the proposed 4D decomposed hash encoding without the unsuitable low-rank assumption and high space complexity.
Additionally, the novel directional attention module effectively aggregates the spatial and temporal features for more accurate deformation prediction across different scene components.
Moreover, we employed smooth regularization to mitigate chaos in deformation prediction, resulting in high rendering quality.
Our experiments demonstrate that Grid4D achieves state-of-the-art performance and delivers high rendering speed for dynamic scene rendering.
However, Grid4D has no improvement in training speed, and like the other dynamic scene rendering models, Grid4D might have artifacts when facing several dynamic scenes with complex and large motions.
Addressing these challenges remains an area for future research.


{
    \small
    \bibliographystyle{plain}
    \bibliography{references}

\begin{thebibliography}{10}

\bibitem{ndr}
Hongrui Cai, Wanquan Feng, Xuetao Feng, Yan Wang, and Juyong Zhang.
\newblock Neural surface reconstruction of dynamic scenes with monocular {RGB-D} camera.
\newblock {\em Advances in Neural Information Processing Systems}, 35:967--981, 2022.

\bibitem{hexplane}
Ang Cao and Justin Johnson.
\newblock {HexPlane}: A fast representation for dynamic scenes.
\newblock In {\em Proceedings of the IEEE/CVF Conference on Computer Vision and Pattern Recognition}, pages 130--141, 2023.

\bibitem{eg3d}
Eric~R. Chan, Connor~Z. Lin, Matthew~A. Chan, Koki Nagano, Boxiao Pan, Shalini~De Mello, Orazio Gallo, Leonidas Guibas, Jonathan Tremblay, Sameh Khamis, Tero Karras, and Gordon Wetzstein.
\newblock Efficient geometry-aware {3D} generative adversarial networks.
\newblock In {\em arXiv}, 2021.

\bibitem{4dregsdf}
Jaesung Choe, Christopher Choy, Jaesik Park, In~So Kweon, and Anima Anandkumar.
\newblock Spacetime surface regularization for neural dynamic scene reconstruction.
\newblock In {\em Proceedings of the IEEE/CVF International Conference on Computer Vision}, pages 17871--17881, 2023.

\bibitem{nerflow}
Yilun Du, Yinan Zhang, Hong-Xing Yu, Joshua~B Tenenbaum, and Jiajun Wu.
\newblock Neural radiance flow for {4D} view synthesis and video processing.
\newblock In {\em 2021 IEEE/CVF International Conference on Computer Vision (ICCV)}, pages 14304--14314. IEEE Computer Society, 2021.

\bibitem{beida4dgs}
Yuanxing Duan, Fangyin Wei, Qiyu Dai, Yuhang He, Wenzheng Chen, and Baoquan Chen.
\newblock {4D Gaussian Splatting}: Towards efficient novel view synthesis for dynamic scenes.
\newblock {\em arXiv preprint arXiv:2402.03307}, 2024.

\bibitem{md-splatting}
Bardienus~P Duisterhof, Zhao Mandi, Yunchao Yao, Jia-Wei Liu, Mike~Zheng Shou, Shuran Song, and Jeffrey Ichnowski.
\newblock {MD-Splatting}: Learning metric deformation from {4D} {Gaussians} in highly deformable scenes.
\newblock {\em arXiv preprint arXiv:2312.00583}, 2023.

\bibitem{tineuvox}
Jiemin Fang, Taoran Yi, Xinggang Wang, Lingxi Xie, Xiaopeng Zhang, Wenyu Liu, Matthias Nie{\ss}ner, and Qi~Tian.
\newblock Fast dynamic radiance fields with time-aware neural voxels.
\newblock In {\em SIGGRAPH Asia 2022 Conference Papers}, pages 1--9, 2022.

\bibitem{k-planes}
Sara Fridovich-Keil, Giacomo Meanti, Frederik~Rahb{\ae}k Warburg, Benjamin Recht, and Angjoo Kanazawa.
\newblock {K-Planes}: Explicit radiance fields in space, time, and appearance.
\newblock In {\em Proceedings of the IEEE/CVF Conference on Computer Vision and Pattern Recognition}, pages 12479--12488, 2023.

\bibitem{dynamic-nerf}
Chen Gao, Ayush Saraf, Johannes Kopf, and Jia-Bin Huang.
\newblock Dynamic view synthesis from dynamic monocular video.
\newblock In {\em Proceedings of the IEEE/CVF International Conference on Computer Vision}, pages 5712--5721, 2021.

\bibitem{forwardflowdnerf}
Xiang Guo, Jiadai Sun, Yuchao Dai, Guanying Chen, Xiaoqing Ye, Xiao Tan, Errui Ding, Yumeng Zhang, and Jingdong Wang.
\newblock Forward flow for novel view synthesis of dynamic scenes.
\newblock In {\em Proceedings of the IEEE/CVF International Conference on Computer Vision}, pages 16022--16033, 2023.

\bibitem{sc-gs}
Yi-Hua Huang, Yang-Tian Sun, Ziyi Yang, Xiaoyang Lyu, Yan-Pei Cao, and Xiaojuan Qi.
\newblock {SC-GS}: Sparse-controlled {Gaussian} splatting for editable dynamic scenes.
\newblock {\em arXiv preprint arXiv:2312.14937}, 2023.

\bibitem{gaussian-splatting}
Bernhard Kerbl, Georgios Kopanas, Thomas Leimk{\"u}hler, and George Drettakis.
\newblock {3D} {Gaussian} splatting for real-time radiance field rendering.
\newblock {\em ACM Transactions on Graphics}, 42(4), 2023.

\bibitem{adam}
Diederik~P Kingma and Jimmy Ba.
\newblock Adam: A method for stochastic optimization.
\newblock {\em arXiv preprint arXiv:1412.6980}, 2014.

\bibitem{dynmf}
Agelos Kratimenos, Jiahui Lei, and Kostas Daniilidis.
\newblock {DynMF}: Neural motion factorization for real-time dynamic view synthesis with {3D} {Gaussian} splatting.
\newblock {\em arXiv preprint arXiv:2312.00112}, 2023.

\bibitem{dynerf}
Tianye Li, Mira Slavcheva, Michael Zollhoefer, Simon Green, Christoph Lassner, Changil Kim, Tanner Schmidt, Steven Lovegrove, Michael Goesele, Richard Newcombe, et~al.
\newblock Neural {3D} video synthesis from multi-view video.
\newblock In {\em Proceedings of the IEEE/CVF Conference on Computer Vision and Pattern Recognition}, pages 5521--5531, 2022.

\bibitem{spacetime-gaussians}
Zhan Li, Zhang Chen, Zhong Li, and Yi~Xu.
\newblock Spacetime {Gaussian} feature splatting for real-time dynamic view synthesis.
\newblock {\em arXiv preprint arXiv:2312.16812}, 2023.

\bibitem{li2021neural}
Zhengqi Li, Simon Niklaus, Noah Snavely, and Oliver Wang.
\newblock Neural scene flow fields for space-time view synthesis of dynamic scenes.
\newblock In {\em Proceedings of the IEEE/CVF Conference on Computer Vision and Pattern Recognition}, pages 6498--6508, 2021.

\bibitem{dynibar}
Zhengqi Li, Qianqian Wang, Forrester Cole, Richard Tucker, and Noah Snavely.
\newblock {DynIBaR}: Neural dynamic image-based rendering.
\newblock In {\em Proceedings of the IEEE/CVF Conference on Computer Vision and Pattern Recognition}, pages 4273--4284, 2023.

\bibitem{gaufre}
Yiqing Liang, Numair Khan, Zhengqin Li, Thu Nguyen-Phuoc, Douglas Lanman, James Tompkin, and Lei Xiao.
\newblock {GauFRe}: {Gaussian} deformation fields for real-time dynamic novel view synthesis.
\newblock {\em arXiv preprint arXiv:2312.11458}, 2023.

\bibitem{saff}
Yiqing Liang, Eliot Laidlaw, Alexander Meyerowitz, Srinath Sridhar, and James Tompkin.
\newblock Semantic attention flow fields for monocular dynamic scene decomposition.
\newblock In {\em Proceedings of the IEEE/CVF International Conference on Computer Vision}, pages 21797--21806, 2023.

\bibitem{gaussian-flow}
Youtian Lin, Zuozhuo Dai, Siyu Zhu, and Yao Yao.
\newblock {Gaussian-Flow}: {4D} reconstruction with dynamic {3D Gaussian} particle.
\newblock {\em arXiv preprint arXiv:2312.03431}, 2023.

\bibitem{rodynrf}
Yu-Lun Liu, Chen Gao, Andreas Meuleman, Hung-Yu Tseng, Ayush Saraf, Changil Kim, Yung-Yu Chuang, Johannes Kopf, and Jia-Bin Huang.
\newblock Robust dynamic radiance fields.
\newblock In {\em Proceedings of the IEEE/CVF Conference on Computer Vision and Pattern Recognition}, pages 13--23, 2023.

\bibitem{gags}
Zhicheng Lu, Xiang Guo, Le~Hui, Tianrui Chen, Min Yang, Xiao Tang, Feng Zhu, and Yuchao Dai.
\newblock {3D} geometry-aware deformable {Gaussian} splatting for dynamic view synthesis.
\newblock {\em arXiv preprint arXiv:2404.06270}, 2024.

\bibitem{dynamic-3d-gaussians}
Jonathon Luiten, Georgios Kopanas, Bastian Leibe, and Deva Ramanan.
\newblock {Dynamic 3D Gaussians}: Tracking by persistent dynamic view synthesis.
\newblock {\em arXiv preprint arXiv:2308.09713}, 2023.

\bibitem{nerf}
Ben Mildenhall, Pratul~P Srinivasan, Matthew Tancik, Jonathan~T Barron, Ravi Ramamoorthi, and Ren Ng.
\newblock {NeRF}: Representing scenes as neural radiance fields for view synthesis.
\newblock {\em Communications of the ACM}, 65(1):99--106, 2021.

\bibitem{instant-ngp}
Thomas M{\"u}ller, Alex Evans, Christoph Schied, and Alexander Keller.
\newblock Instant neural graphics primitives with a multiresolution hash encoding.
\newblock {\em ACM Transactions on Graphics (ToG)}, 41(4):1--15, 2022.

\bibitem{hypernerf}
Keunhong Park, Utkarsh Sinha, Peter Hedman, Jonathan~T Barron, Sofien Bouaziz, Dan~B Goldman, Ricardo Martin-Brualla, and Steven~M Seitz.
\newblock {HyperNeRF}: A higher-dimensional representation for topologically varying neural radiance fields.
\newblock {\em arXiv preprint arXiv:2106.13228}, 2021.

\bibitem{park2023temporal}
Sungheon Park, Minjung Son, Seokhwan Jang, Young~Chun Ahn, Ji-Yeon Kim, and Nahyup Kang.
\newblock Temporal interpolation is all you need for dynamic neural radiance fields.
\newblock In {\em Proceedings of the IEEE/CVF Conference on Computer Vision and Pattern Recognition}, pages 4212--4221, 2023.

\bibitem{pytorch}
Adam Paszke, Sam Gross, Francisco Massa, Adam Lerer, James Bradbury, Gregory Chanan, Trevor Killeen, Zeming Lin, Natalia Gimelshein, Luca Antiga, et~al.
\newblock {PyTorch}: An imperative style, high-performance deep learning library.
\newblock {\em Advances in neural information processing systems}, 32, 2019.

\bibitem{d-nerf}
Albert Pumarola, Enric Corona, Gerard Pons-Moll, and Francesc Moreno-Noguer.
\newblock {D-NeRF}: Neural radiance fields for dynamic scenes.
\newblock In {\em Proceedings of the IEEE/CVF Conference on Computer Vision and Pattern Recognition}, pages 10318--10327, 2021.

\bibitem{colmap}
Johannes~L Schonberger and Jan-Michael Frahm.
\newblock {Structure-from-Motion} revisited.
\newblock In {\em Proceedings of the IEEE conference on computer vision and pattern recognition}, pages 4104--4113, 2016.

\bibitem{tensor4d}
Ruizhi Shao, Zerong Zheng, Hanzhang Tu, Boning Liu, Hongwen Zhang, and Yebin Liu.
\newblock {Tensor4D}: Efficient neural {4D} decomposition for high-fidelity dynamic reconstruction and rendering.
\newblock In {\em Proceedings of the IEEE/CVF Conference on Computer Vision and Pattern Recognition}, pages 16632--16642, 2023.

\bibitem{vgg}
Karen Simonyan and Andrew Zisserman.
\newblock Very deep convolutional networks for large-scale image recognition.
\newblock {\em arXiv preprint arXiv:1409.1556}, 2014.

\bibitem{factorized-motion}
Nagabhushan Somraj, Kapil Choudhary, Sai~Harsha Mupparaju, and Rajiv Soundararajan.
\newblock Factorized motion fields for fast sparse input dynamic view synthesis.
\newblock {\em arXiv preprint arXiv:2404.11669}, 2024.

\bibitem{nerfplayer}
Liangchen Song, Anpei Chen, Zhong Li, Zhang Chen, Lele Chen, Junsong Yuan, Yi~Xu, and Andreas Geiger.
\newblock {NeRFPlayer}: A streamable dynamic scene representation with decomposed neural radiance fields.
\newblock {\em IEEE Transactions on Visualization and Computer Graphics}, 29(5):2732--2742, 2023.

\bibitem{3dgstream}
Jiakai Sun, Han Jiao, Guangyuan Li, Zhanjie Zhang, Lei Zhao, and Wei Xing.
\newblock {3DGStream}: On-the-fly training of {3D Gaussians} for efficient streaming of photo-realistic free-viewpoint videos.
\newblock {\em arXiv preprint arXiv:2403.01444}, 2024.

\bibitem{nr-nerf}
Edgar Tretschk, Ayush Tewari, Vladislav Golyanik, Michael Zollh{\"o}fer, Christoph Lassner, and Christian Theobalt.
\newblock {Non-Rigid Neural Radiance Fields}: Reconstruction and novel view synthesis of a dynamic scene from monocular video.
\newblock In {\em Proceedings of the IEEE/CVF International Conference on Computer Vision}, pages 12959--12970, 2021.

\bibitem{t-sne}
Laurens Van~der Maaten and Geoffrey Hinton.
\newblock Visualizing data using {t-SNE}.
\newblock {\em Journal of machine learning research}, 9(11), 2008.

\bibitem{fsdnerf}
Chaoyang Wang, Lachlan~Ewen MacDonald, Laszlo~A Jeni, and Simon Lucey.
\newblock Flow supervision for deformable {NeRF}.
\newblock In {\em Proceedings of the IEEE/CVF Conference on Computer Vision and Pattern Recognition}, pages 21128--21137, 2023.

\bibitem{masked-spacetime-hashing}
Feng Wang, Zilong Chen, Guokang Wang, Yafei Song, and Huaping Liu.
\newblock Masked space-time hash encoding for efficient dynamic scene reconstruction.
\newblock {\em Advances in Neural Information Processing Systems}, 2023.

\bibitem{mixvoxels}
Feng Wang, Sinan Tan, Xinghang Li, Zeyue Tian, Yafei Song, and Huaping Liu.
\newblock Mixed neural voxels for fast multi-view video synthesis.
\newblock In {\em Proceedings of the IEEE/CVF International Conference on Computer Vision}, pages 19706--19716, 2023.

\bibitem{ssim}
Zhou Wang, Alan~C Bovik, Hamid~R Sheikh, and Eero~P Simoncelli.
\newblock Image quality assessment: from error visibility to structural similarity.
\newblock {\em IEEE transactions on image processing}, 13(4):600--612, 2004.

\bibitem{4d-gs}
Guanjun Wu, Taoran Yi, Jiemin Fang, Lingxi Xie, Xiaopeng Zhang, Wei Wei, Wenyu Liu, Qi~Tian, and Xinggang Wang.
\newblock {4D} {Gaussian} splatting for real-time dynamic scene rendering.
\newblock {\em arXiv preprint arXiv:2310.08528}, 2023.

\bibitem{s-nerf}
Ziyang Xie, Junge Zhang, Wenye Li, Feihu Zhang, and Li~Zhang.
\newblock {S-NeRF}: Neural radiance fields for street views.
\newblock {\em arXiv preprint arXiv:2303.00749}, 2023.

\bibitem{4k4d}
Zhen Xu, Sida Peng, Haotong Lin, Guangzhao He, Jiaming Sun, Yujun Shen, Hujun Bao, and Xiaowei Zhou.
\newblock {4K4D}: Real-time {4D} view synthesis at {4K} resolution.
\newblock In {\em Proceedings of the IEEE/CVF Conference on Computer Vision and Pattern Recognition}, pages 20029--20040, 2024.

\bibitem{nerf-ds}
Zhiwen Yan, Chen Li, and Gim~Hee Lee.
\newblock {NeRF-DS}: Neural radiance fields for dynamic specular objects.
\newblock In {\em Proceedings of the IEEE/CVF Conference on Computer Vision and Pattern Recognition}, pages 8285--8295, 2023.

\bibitem{banmo}
Gengshan Yang, Minh Vo, Natalia Neverova, Deva Ramanan, Andrea Vedaldi, and Hanbyul Joo.
\newblock {BANMo}: Building animatable {3D} neural models from many casual videos.
\newblock In {\em Proceedings of the IEEE/CVF Conference on Computer Vision and Pattern Recognition}, pages 2863--2873, 2022.

\bibitem{fudan4dgs}
Zeyu Yang, Hongye Yang, Zijie Pan, Xiatian Zhu, and Li~Zhang.
\newblock Real-time photorealistic dynamic scene representation and rendering with {4D} {Gaussian} splatting.
\newblock {\em arXiv preprint arXiv:2310.10642}, 2023.

\bibitem{deformable-3d-gaussians}
Ziyi Yang, Xinyu Gao, Wen Zhou, Shaohui Jiao, Yuqing Zhang, and Xiaogang Jin.
\newblock Deformable {3D} {Gaussians} for high-fidelity monocular dynamic scene reconstruction.
\newblock {\em arXiv preprint arXiv:2309.13101}, 2023.

\bibitem{differentiable-splatting}
Wang Yifan, Felice Serena, Shihao Wu, Cengiz {\"O}ztireli, and Olga Sorkine-Hornung.
\newblock Differentiable surface splatting for point-based geometry processing.
\newblock {\em ACM Transactions on Graphics (TOG)}, 38(6):1--14, 2019.

\bibitem{cogs}
Heng Yu, Joel Julin, Zolt{\'a}n~{\'A} Milacski, Koichiro Niinuma, and L{\'a}szl{\'o}~A Jeni.
\newblock {CoGS}: Controllable {Gaussian} splatting.
\newblock {\em arXiv preprint arXiv:2312.05664}, 2023.

\bibitem{star}
Wentao Yuan, Zhaoyang Lv, Tanner Schmidt, and Steven Lovegrove.
\newblock {STaR}: Self-supervised tracking and reconstruction of rigid objects in motion with neural rendering.
\newblock In {\em Proceedings of the IEEE/CVF Conference on Computer Vision and Pattern Recognition}, pages 13144--13152, 2021.

\bibitem{lpips}
Richard Zhang, Phillip Isola, Alexei~A Efros, Eli Shechtman, and Oliver Wang.
\newblock The unreasonable effectiveness of deep features as a perceptual metric.
\newblock In {\em Proceedings of the IEEE conference on computer vision and pattern recognition}, pages 586--595, 2018.

\bibitem{ndf}
Ruiqi Zhang and Jie Chen.
\newblock {NDF}: Neural deformable fields for dynamic human modelling.
\newblock In {\em European Conference on Computer Vision}, pages 37--52. Springer, 2022.

\bibitem{splatting}
Matthias Zwicker, Hanspeter Pfister, Jeroen Van~Baar, and Markus Gross.
\newblock Surface splatting.
\newblock In {\em Proceedings of the 28th annual conference on Computer graphics and interactive techniques}, pages 371--378, 2001.

\end{thebibliography}
}

\clearpage
\appendix

\section*{\centering \Large --~Supplementary~--}
\section{Details of Experimental Setup}\label{sec:supp-setup}

\begin{figure}[hp]
  \centering
  \includegraphics[scale=0.28, trim=0 330 350 0, clip]{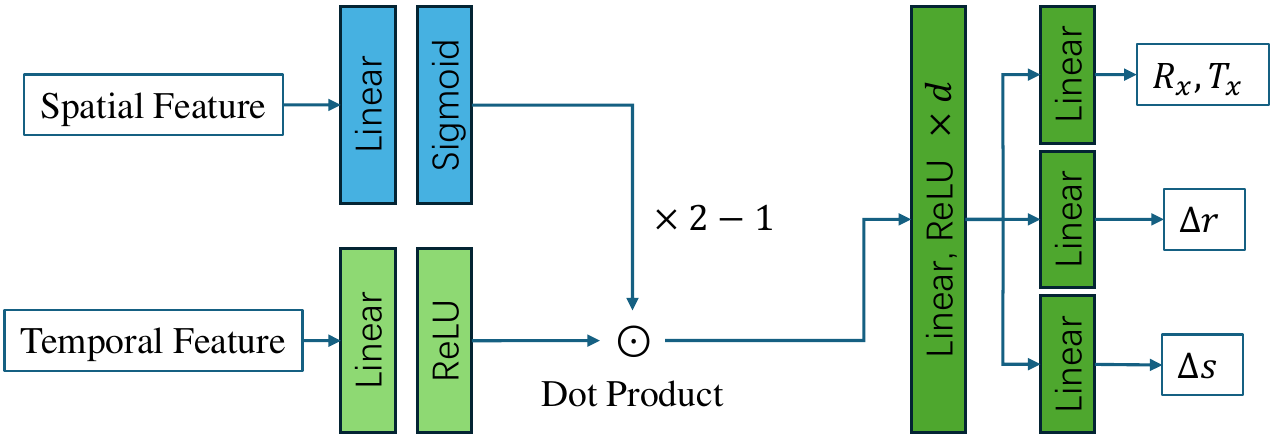}
  \caption{Architecture of our multi-head directional attention decoder.}
  \label{fig:supp-network}
\end{figure}

\textbf{Network Architecture of Grid4D.}
The upper limit of the hash table size is $2^{19}$ for both spatial and temporal 3D grids.
The feature dimension of each voxel is 2 for all grids.
We set the spatial dimension resolutions of temporal grids according to the scale of the scene.
We usually set the time dimension resolution to a value between a half and a quarter of the time samples.
The architecture of our multi-head directional attention decoder is illustrated by Figure~\ref{fig:supp-network}.
The spatial and temporal MLPs only have one fully connected layer and one activation layer.
For the multi-head deformation decoder, we set the depth to two for the HyperNeRF~\cite{hypernerf} dataset and one for the D-NeRF~\cite{d-nerf} dataset, including the output layer, and set all the widths to 256.

\textbf{Optimization.}
The scheduler of the learning rate primarily follows DeformGS~\cite{deformable-3d-gaussians, gaussian-splatting}.
The loss weight~$\lambda_c$ and $\lambda_r$ in Equation~\ref{equa:total-loss} is set to 0.2 and 0.5 for common scenes.
Notably, the learning rate of the MLP decoder is determined based on the scale of the scene.
Additionally, the learning rate of the grid hash encoders is set to 10\textasciitilde 50 times larger than the MLP decoder.
We use Adam~\cite{adam} optimizer with $\beta=(0.9, 0.999)$ for training and set the background to black.
Due to the differences between the spatial and temporal grids, we set different smooth regularization parameters for the $(x, y, z)$ grid in several scenes.
For the scenes in the HyperNeRF~\cite{hypernerf} and Neu3D~\cite{dynerf} dataset, we use the SfM~\cite{colmap} points to initialize Gaussians.

\textbf{Deformation Map.}
The formula for deformation maps is
\begin{align}
\Delta x = \frac{f(x, t + \tau) - f(x, t)}{\tau}
\end{align}
where $x$ is the canonical Gaussian position, and $f$ is the deformation field.
For all experiments, We set $\tau$ to 0.05, and limit the absolute value of $\Delta x$ for color.
This map shows the situation of predicted deformation: the similar colors of two parts mean similar deformation sizes on each axis.

\section{Additional Comparisons}\label{sec:supp-comparison}

\begin{figure}[tb]
  \centering
  \includegraphics[scale=0.88, trim=0 150 230 0, clip]{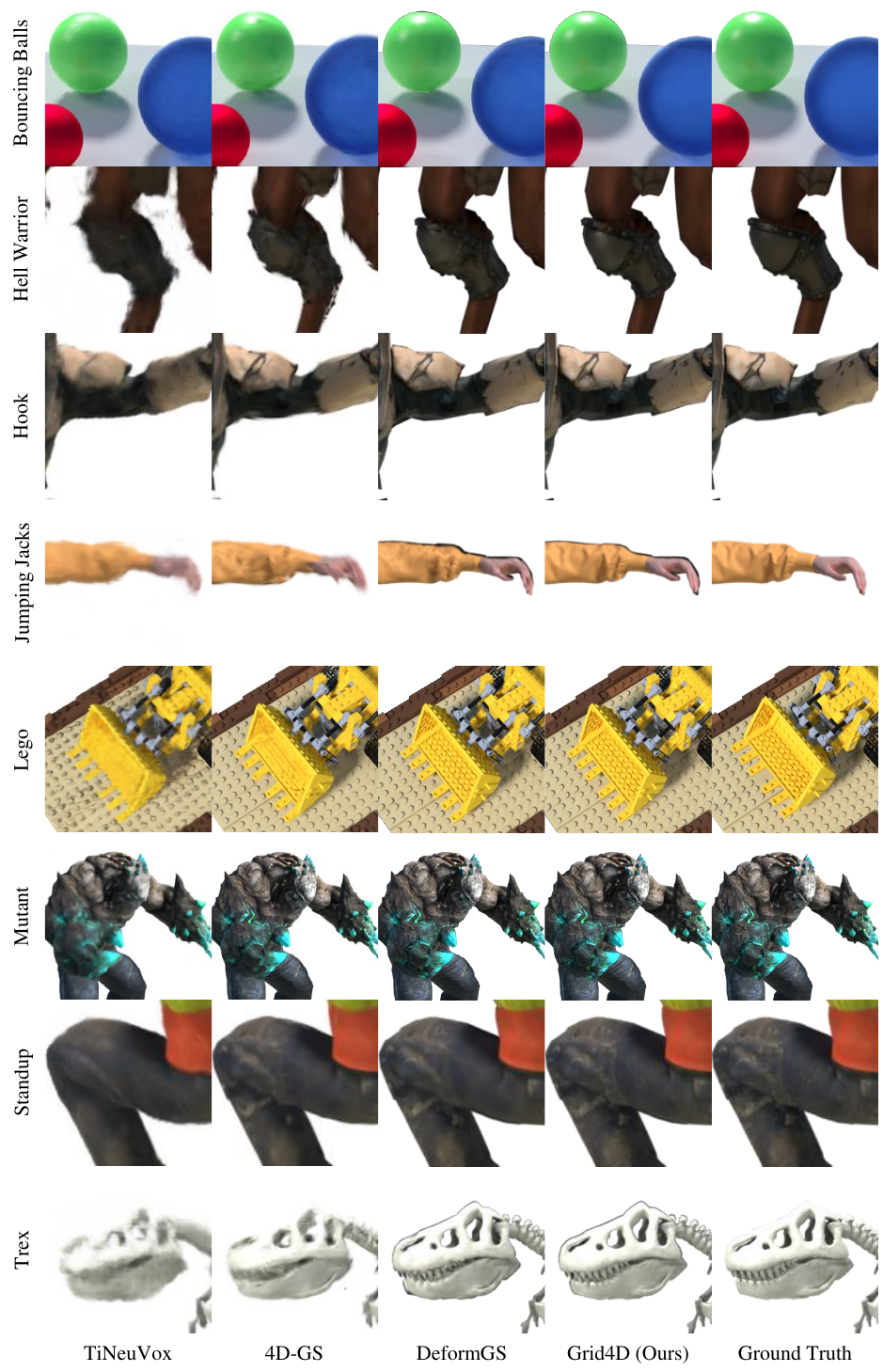}
  \caption{Additional qualitative comparisons on the synthetic D-NeRF~\cite{d-nerf} dataset.}
  \label{fig:supp-d-nerf}
\end{figure}

\textbf{Additional Results on D-NeRF~\cite{d-nerf} Dataset.}
We visualize more experimental results on the D-NeRF~\cite{d-nerf} dataset in Figure~\ref{fig:supp-d-nerf}.
We observe that Grid4D exhibits superior rendering quality compared to the state-of-the-art models.

\begin{figure}[!tb]
  \centering
  \includegraphics[scale=0.68, trim=0 260 110 0, clip]{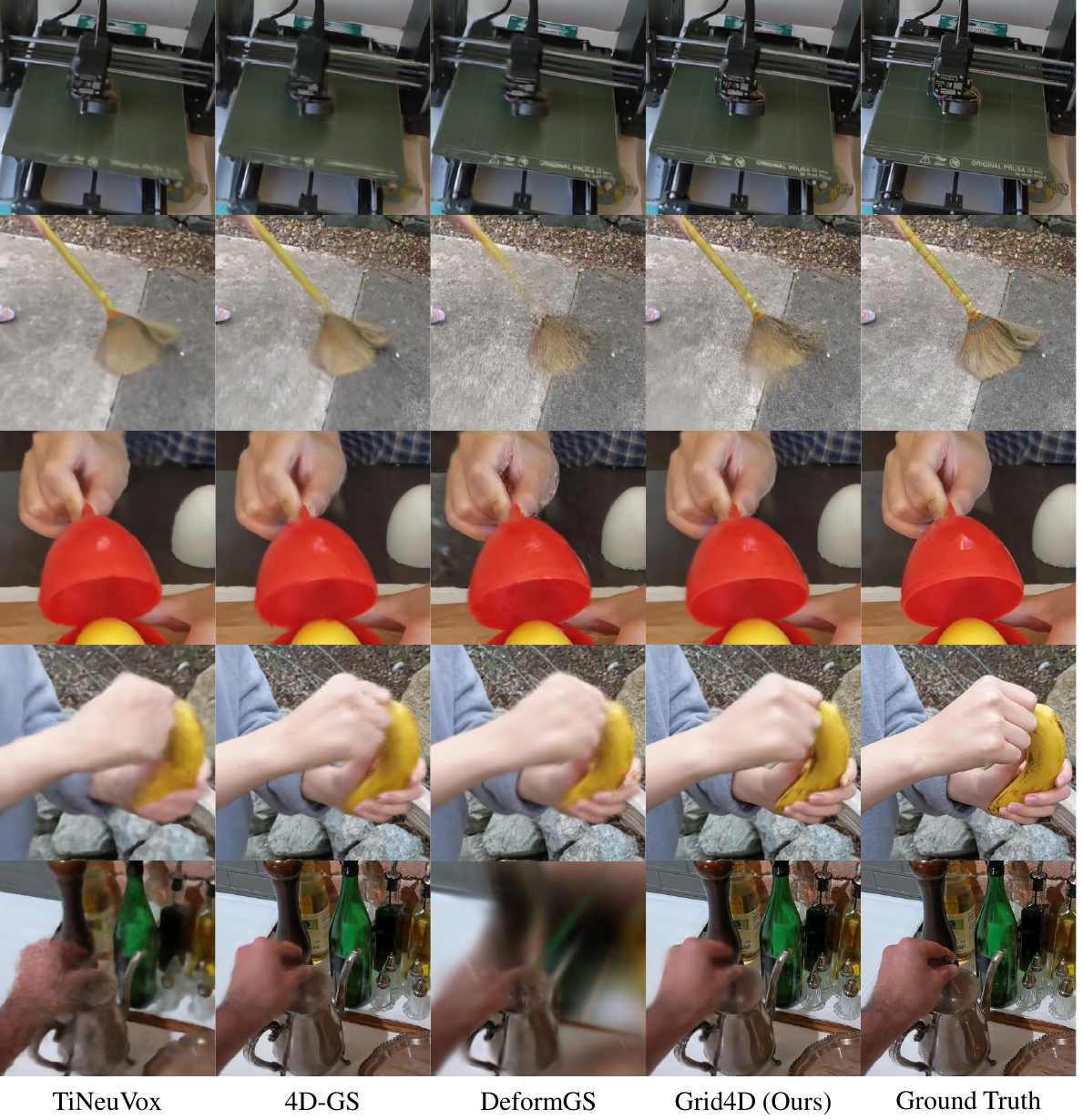}
  \caption{Additional qualitative comparisons on the real-world HyperNeRF~\cite{hypernerf} dataset.}
  \label{fig:supp-hypernerf1}
\end{figure}

\begin{figure}[!tb]
  \centering
  \includegraphics[scale=0.68, trim=0 260 110 0, clip]{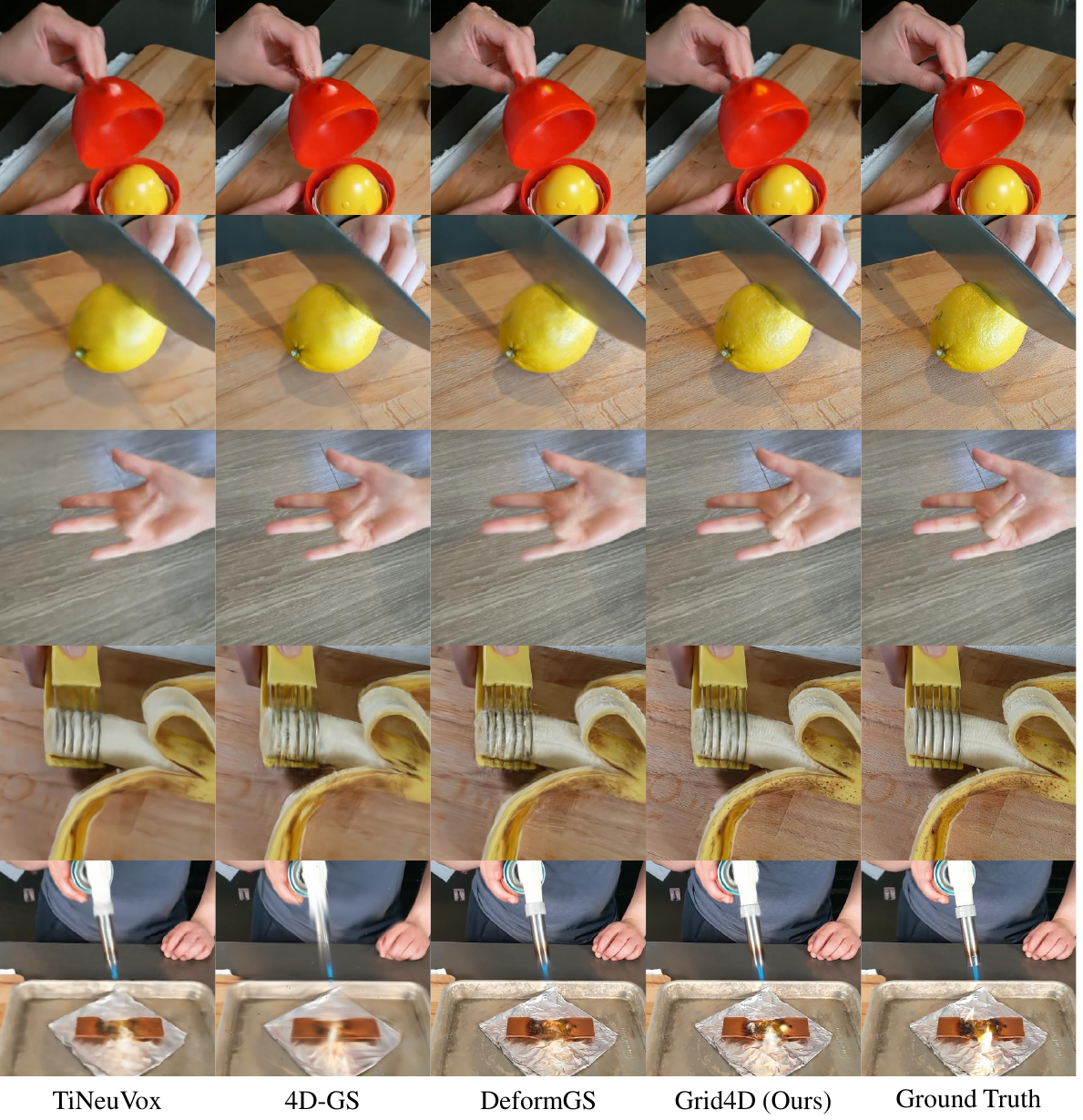}
  \caption{Additional qualitative comparisons on the real-world HyperNeRF~\cite{hypernerf} dataset.}
  \label{fig:supp-hypernerf2}
\end{figure}

\textbf{Per Scene Results on HyperNeRF~\cite{hypernerf} Dataset.}
We provide the per-scene results for the experiments on the real-world HyperNeRF~\cite{hypernerf} dataset.
Table~\ref{tab:supp-hypernerf}, Figure~\ref{fig:supp-hypernerf1} and Figure~\ref{fig:supp-hypernerf2} illustrate the comparisons.
While the quantitative results for Grid4D do not surpass those of other models in several scenes, it is noteworthy that our model exhibits significantly improved clarity in rendering, as demonstrated in Figure~\ref{fig:supp-hypernerf1} and Figure~\ref{fig:supp-hypernerf2}.
We also find reconstruction failures of DeformGS~\cite{deformable-3d-gaussians} in the `Teapot' and `Broom' scenes (the second and last line of Figure~\ref{fig:supp-hypernerf1}), as mentioned in their paper.

\section{Additional Ablations}\label{sec:supp-ablation}

\textbf{Additional Architecture Ablations.}
We also conduct more ablation studies for Grid4D.
We change the depth $d$ of the multi-head decoder to one and two for the D-NeRF dataset, and the max level number $L$ of the temporal grid hash encoder to 8 and 16.
Additionally, we apply the simple position deformation method to Grid4D as Grid4D w/o RT, which directly adds the deformation to the position and is used in the prior works~\cite{4d-gs, deformable-3d-gaussians}.
The results can be found in the Table~\ref{tab:supp-ablation-d-nerf}.
We can find that when the model becomes deeper, the performance might become worse, and we consider that the reason might be the training difficulties of deep MLPs.

\begin{table}[tb]
\caption{Per scene comparisons on the real world HyperNeRF~\cite{hypernerf} `vrig' and `interp' dataset. The higher PSNR~$(\uparrow)$ and higher MS-SSIM~$\uparrow$ denote better rendering quality. The color of each cell shows the \sethlcolor{pink!150}\hl{best} and the \sethlcolor{orange!40}\hl{second best}. We set all rendering resolutions to $536\times 900$.}
\label{tab:supp-hypernerf}
\centering
\small
\setlength{\tabcolsep}{0.7mm}{
\begin{tabular}{ccccccccccc}
\toprule
         & \multicolumn{2}{c}{3D Printer} & \multicolumn{2}{c}{Broom} & \multicolumn{2}{c}{Chicken~(vrig)} & \multicolumn{2}{c}{Peel Banana} & \multicolumn{2}{c}{Teapot} \\
Model                    & PSNR & MS-SSIM & PSNR & MS-SSIM & PSNR & MS-SSIM & PSNR & MS-SSIM & PSNR & MS-SSIM \\
\midrule
TiNeuVox~\cite{tineuvox} & \cellcolor{pink!150} 22.77 & \cellcolor{pink!150} 0.839 & 21.27 & 0.683 & 28.27 & \cellcolor{pink!150} 0.948 & 24.50 & 0.874 & 24.15 & 0.893 \\
4D-GS~\cite{4d-gs} & 22.00 & 0.807 & \cellcolor{orange!40} 21.80 & \cellcolor{orange!40} 0.684 & \cellcolor{orange!40} 28.55 & 0.927 & \cellcolor{orange!40} 27.62 & \cellcolor{orange!40} 0.934 & \cellcolor{pink!150} 26.99 & \cellcolor{pink!150} 0.941\\
Grid4D~(Ours) & \cellcolor{orange!40} 22.35 & \cellcolor{orange!40} 0.825 & \cellcolor{pink!150} 21.86 & \cellcolor{pink!150} 0.710 & \cellcolor{pink!150} 29.26 & \cellcolor{orange!40} 0.942 & \cellcolor{pink!150} 28.53 & \cellcolor{pink!150} 0.946 & \cellcolor{orange!40} 26.53 & \cellcolor{orange!40} 0.932 \\
\toprule
         & \multicolumn{2}{c}{Chicken~(interp)} & \multicolumn{2}{c}{Cut Lemon} & \multicolumn{2}{c}{Hand} & \multicolumn{2}{c}{Slice Banana} & \multicolumn{2}{c}{Chocolate} \\
Model                    & PSNR & MS-SSIM & PSNR & MS-SSIM & PSNR & MS-SSIM & PSNR & MS-SSIM & PSNR & MS-SSIM \\
\midrule
TiNeuVox~\cite{tineuvox} & \cellcolor{pink!150} 27.69 & \cellcolor{pink!150} 0.951 & 28.54 & \cellcolor{orange!40} 0.955 & 27.44 & 0.872 & \cellcolor{pink!150} 27.67 & \cellcolor{pink!150} 0.916 & \cellcolor{orange!40} 26.97 & \cellcolor{orange!40} 0.948 \\
4D-GS~\cite{4d-gs} & 26.91 & 0.912 & \cellcolor{orange!40} 30.26 & 0.936 & \cellcolor{orange!40} 29.87 & \cellcolor{orange!40} 0.939 & 25.18 & 0.812 & 26.05 & 0.933 \\
Grid4D~(Ours) & \cellcolor{orange!40} 27.31 & \cellcolor{orange!40} 0.924 & \cellcolor{pink!150} 32.18 & \cellcolor{pink!150} 0.967 & \cellcolor{pink!150} 31.31 & \cellcolor{pink!150} 0.958 & \cellcolor{orange!40} 25.79 & \cellcolor{orange!40} 0.853 & \cellcolor{pink!150} 28.23 & \cellcolor{pink!150} 0.965 \\
\bottomrule
\end{tabular}}
\end{table}

\begin{table}[!tb]
\caption{Additional ablation results of model architecture on the synthetic D-NeRF~\cite{d-nerf} dataset. The color of each cell shows the \sethlcolor{pink!150}\hl{best} and the \sethlcolor{orange!40}\hl{second best}.}
\label{tab:supp-ablation-d-nerf}
\centering
\small
\setlength{\tabcolsep}{1.1mm}{
\begin{tabular}{ccccccc}
\toprule
Model                    & $L=8, d=0$ & $L=16, d=0$ & $L=32, d=0$~(Ours) & $L=32, d=1$ & $L=32, d=2$ & w/o RT\\
\midrule
PSNR~($\uparrow$) & 41.34 & 41.69 & \cellcolor{pink!150} 42.00 & \cellcolor{orange!40} 41.96 & 41.75 & 41.80 \\
\bottomrule
\end{tabular}}
\end{table}

\begin{table}[!tb]
\caption{Comparison of average training computational cost on the D-NeRF~\cite{d-nerf} dataset with a single RTX 3090 GPU.}
\label{tab:supp-speed}
\centering
\small
\setlength{\tabcolsep}{3mm}{
\begin{tabular}{ccccc}
\toprule
Model & 4D-GS~\cite{4d-gs} & DeformGS~\cite{deformable-3d-gaussians} & SC-GS~\cite{sc-gs} & Grid4D~(Ours) \\
\midrule
Time & 20min & 33min & 75min & 55min \\
GPU Memory & 1GB & 4.5GB & 3.1GB & 4.0GB \\
PSNR & 35.41 & 40.16 & 41.65 & 42.00 \\
\bottomrule
\end{tabular}}
\end{table}

\textbf{Additional Smooth Regularization Ablations.}
We conduct a smooth regularization for the both grid hash encoder and MLP decoder in Grid4D w both by adding the following loss,
\begin{align}
    L_d = ||R_x - R_{x + \epsilon}||_2^2 + ||T_x - T_{x + \epsilon}||_2^2
\end{align}
where $R_x, T_x$ are the predicted position deformation of the 4D input $x$, and $R_{x + \epsilon}, T_{x + \epsilon}$ are the deformation of the perturbed 4D input $x + \epsilon$.
The results are shown in the left part of Table~\ref{tab:supp-reg}, and we can find that the smooth regularization of the MLP decoder does not make sense.
We consider that this is because of the smooth inherent property of MLPs.

\begin{figure}[!tb]
  \centering
  \includegraphics[scale=0.5, trim=0 490 0 0, clip]{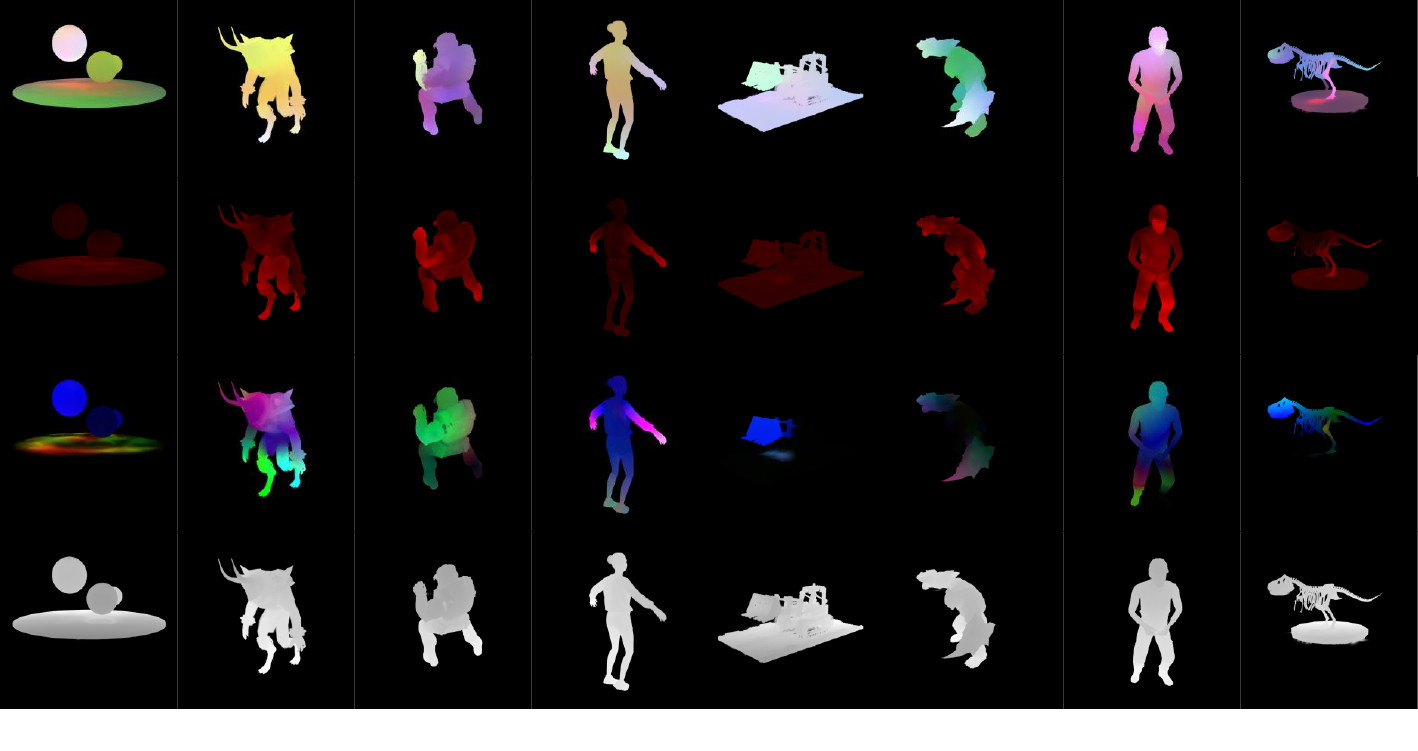}
  \caption{Additional analysis of Grid4D. Each line in turn shows the temporal feature maps, spatial feature maps, deformation maps, and depth maps. The lighter parts in the feature maps denote the stronger activation of the corresponding features.}
  \label{fig:supp-analysis}
\end{figure}

\textbf{Visualization of Feature, Deformation, and Depth Maps.}
We also visualize the feature maps by projecting the L2 norm of the features encoding the Gaussian positions from our 4D decomposed hash encoder.
We set the RGB color of the temporal feature map to the L2 norm of $(x, y, t), (y, z, t), (x, z, t)$ grid features, and the feature map is rendered by the rasterization of Gaussians with the specified color.
The results in Figure~\ref{fig:supp-analysis} denote that the proposed encoding method effectively represents the deformation features in both temporal and spatial spaces.
Also, the depth maps show that we have a precise depth prediction.


\section{Limitation}\label{sec:supp-limitation}
\begin{figure}[!tb]
  \centering
  \includegraphics[scale=0.43, trim=0 400 40 0, clip]{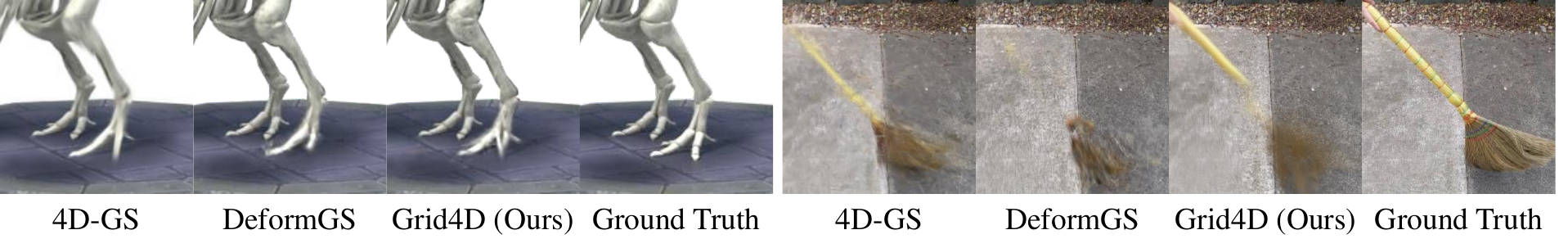}
  \caption{Artifacts of Grid4D and other state-of-the-art models~\cite{4d-gs, deformable-3d-gaussians}.}
  \label{fig:supp-limitation}
\end{figure}

Although our model achieves state-of-the-art performance with our proposed explicit representation, Grid4D has no improvement in training speed.
However, compared to DeformGS~\cite{deformable-3d-gaussians}, Grid4D has less memory overhead.
As shown in Table~\ref{tab:supp-speed}, the computational cost has little influence on model training.
Figure~\ref{fig:supp-limitation} displays several artifacts when Grid4D and the state-of-the-art models~\cite{4d-gs, deformable-3d-gaussians} render several scenes with large and complex motions.

\begin{table}[!tb]
\caption{Additional ablation results of adding MLP decoder regularization on the D-NeRF~\cite{d-nerf} dataset.}
\label{tab:supp-reg}
\centering
\small
\begin{tabular}{ccc}
\toprule
Model & Grid4D w both & Grid4D~(Ours) \\
\midrule
PSNR~($\uparrow$) & 41.92 & 42.00 \\
\bottomrule
\end{tabular}
\end{table}

\end{document}